\documentclass{article}

\usepackage{arxiv}

\usepackage[utf8]{inputenc} % allow utf-8 input
\usepackage[T1]{fontenc}    % use 8-bit T1 fonts
\usepackage{url}            % simple URL typesetting
\usepackage{booktabs}       % professional-quality tables
\usepackage{amsfonts}       % blackboard math symbols
\usepackage{nicefrac}       % compact symbols for 1/2, etc.
\usepackage{microtype}      % microtypography
\usepackage{lipsum}		% Can be removed after putting your text content
\usepackage{graphicx}
\usepackage{subcaption}
\usepackage{booktabs}
\usepackage{mwe}
\usepackage{listings}
\usepackage{caption}
\usepackage{amsmath}
\usepackage{longtable}
\usepackage[CaptionAfterwards]{fltpage}

\DeclareFixedFont{\ttb}{T1}{txtt}{bx}{n}{12} % for bold
\DeclareFixedFont{\ttm}{T1}{txtt}{m}{n}{12}  % for normal

\usepackage{hyperref}

\title{An Uncertainty-Aware, Shareable and Transparent Neural Network Architecture for Brain-Age Modeling}

\date{}

\author{
Tim Hahn\textsuperscript{1}\thanks{
	These authors contributed equally	
}
\And Jan Ernsting\textsuperscript{1,2$\ast$}
\And Nils R. Winter\textsuperscript{1}
\And Vincent Holstein\textsuperscript{1}
\And Ramona Leenings\textsuperscript{1,2}
\And Marie Beisemann\textsuperscript{3}
\And Lukas Fisch\textsuperscript{1}
\And Kelvin Sarink\textsuperscript{1}
\And Daniel Emden\textsuperscript{1}
\And Nils Opel\textsuperscript{1}
\And Ronny Redlich\textsuperscript{1,4}
\And Jonathan Repple\textsuperscript{1}
\And Dominik Grotegerd\textsuperscript{1}
\And Susanne Meinert\textsuperscript{1}
\And Jochen G. Hirsch\textsuperscript{5}
\And Thoralf Niendorf\textsuperscript{6}
\And Beate Endemann\textsuperscript{6}
\And Fabian Bamberg\textsuperscript{7}
\And Thomas Kröncke\textsuperscript{8}
\And Robin Bülow\textsuperscript{9}
\And Henry Völzke\textsuperscript{10}
\And Oyunbileg von Stackelberg\textsuperscript{11}
\And Ramona Felizitas Sowade\textsuperscript{11}
\And Lale Umutlu\textsuperscript{12}
\And Börge Schmidt\textsuperscript{12}
\And Svenja Caspers\textsuperscript{13,14}
\And German National Cohort Study Center Consortium
\And Harald Kugel\textsuperscript{15}
\And Tilo Kircher\textsuperscript{16}
\And Benjamin Risse\textsuperscript{2}
\And Christian Gaser\textsuperscript{17}
\And James H. Cole\textsuperscript{18,19}
\And Udo Dannlowski\textsuperscript{1}
\And Klaus Berger\textsuperscript{20}
}

\renewcommand{\headeright}{}
\renewcommand{\undertitle}{}

\begin{document}
\maketitle

\section*{Affiliations}
	1 Institute for Translational Psychiatry, University of Münster, Germany
\newline
2 Faculty of Mathematics and Computer Science, University of Münster, Germany 
\newline
3 Department of Statistics, TU Dortmund University, Dortmund, Germany
\newline
4 Department of Psychology, University of Halle, Halle, Germany
\newline
5 Fraunhofer MEVIS, Bremen, Germany
\newline
6 Berlin Ultrahigh Field Facility (B.U.F.F.), NAKO imaging site Berlin, Max-Delbrueck Center for Molecular Medicine in the Helmholtz Association, Berlin, Germany
\newline
7 Department of Radiology, Medical Center - University of Freiburg, Faculty of Medicine, University of Freiburg, Freiburg, Germany
\newline
8 Department of Diagnostic and Interventional Radiology, University Hospital Augsburg, Augsburg, Germany
\newline
9 Institute of Diagnostic Radiology and Neuroradiology, University of Greifswald, Greifswald, Germany
\newline
10 Institute for Community Medicine, University Medicine Greifswald, Greifswald, Germany
\newline
11 Department of Diagnostic and Interventional Radiology, University Hospital Heidelberg, Heidelberg, Germany; and Translational Lung Research Center, Member of the German Lung Research Center, Heidelberg, Germany
\newline
12 Institute for Medical Informatics, Biometry and Epidemiology, University of Duisburg-Essen
\newline
13 Institute for Anatomy I, Medical Faculty and University Hospital Düsseldorf, Heinrich Heine University Düsseldorf, 40225 Düsseldorf, Germany
\newline
14 Institute of Neuroscience and Medicine (INM-1), Research Centre Jülich, 52425 Jülich, Germany
\newline
15 Institute of Clinical Radiology, University of Münster, Germany
\newline
16 Department of Psychiatry and Psychotherapy, Phillips University Marburg, Germany
\newline
17 Department of Psychiatry and Psychotherapy, Jena University Hospital; Department of Neurology, Jena University Hospital, Jena, Germany
\newline
18 Department of Neuroimaging, Institute of Psychiatry, Psychology \& Neuroscience, King’s College London, London, UK
\newline
19 Computational, Cognitive and Clinical Neuroimaging Laboratory, Department King’s College, London, United Kingdom
\newline
20 Institute of Epidemiology and Social Medicine, University of Münster, Münster, Germany
\newline

\begin{abstract}
The deviation between chronological age and age predicted from neuroimaging data has been identified as a sensitive risk-marker of cross-disorder brain changes, growing into a cornerstone of biological age-research. However, Machine Learning models underlying the field do not consider uncertainty, thereby confounding results with training data density and variability. Also, existing models are commonly based on homogeneous training sets, often not independently validated, and cannot be shared due to data protection issues. Here, we introduce an uncertainty-aware, shareable, and transparent Monte-Carlo Dropout Composite-Quantile-Regression (MCCQR) Neural Network trained on N=10,691 datasets from the German National Cohort. The MCCQR model provides robust, distribution-free uncertainty quantification in high-dimensional neuroimaging data, achieving lower error rates compared to existing models across ten recruitment centers and in three independent validation samples (N=4,004). In two examples, we demonstrate that it prevents spurious associations and increases power to detect accelerated brain-aging. We make the pre-trained model publicly available.
\end{abstract}

% keywords can be removed
% \keywords{First keyword \and Second keyword \and More}
\section{Introduction}
% Introduce the problem
Though aging is ubiquitous, the rate at which age-associated biological changes in the brain occur differs substantially between individuals. Building on this, the so-called ‘brain-age paradigm’ \cite{cole2017,cole2017} aims to estimate a brain's “biological age”\cite{ludwig1980} and posits that brain-age may serve as a cumulative marker of disease-risk, functional capacity and residual lifespan{cole2020}. In a typical brain-age study, a machine learning model is trained on neuroimaging data – usually whole-brain structural T1-weighted Magnetic Resonance Imaging (MRI) data – to predict chronological age. This trained model is then used to evaluate neuroimaging data from previously unseen individuals and evaluated based on the “brain-age gap” as defined by the difference between predicted and chronological age.

A decade after its inception, this approach has developed into a major component of biological age research with a plethora of publications linking individual differences between chronological and brain-age to genetic, environmental, and demographic characteristics in health and disease (for a comprehensive review, see \cite{franke2019}). For example, accelerated brain-aging – i.e. a higher brain-age compared to chronological age – has been associated with markers of physiological aging (e.g., grip strength, lung function, walking speed), cognitive aging\cite{cole2018}, life risk\cite{Bittner.2021} poor future health outcomes including progression from mild cognitive impairment to dementia\cite{franke2012,gaser2013}, mortality\cite{cole2018}, as well as a range of neurological diseases and psychiatric disorders (reviewed in \cite{cole2019}). 
However, despite its scientific relevance and popularity, brain-age research faces numerous challenges, which hamper further progress as well as the translation of findings into clinical practice. First, the quantity and quality of the neuroimaging data upon which the underlying brain-age models are trained differ widely across studies, with only more recent studies training on more than 100 samples. Given the large number of voxels measured by modern structural MRI and the complexity of the multivariate Machine Learning models used in brain-age modeling, such small training sample sizes may lead to low-performance models with comparatively large errors. While studies drawing on larger training samples exist, models are of limited value as they cover only a certain age range (cf. for example the UK Biobank\cite{cole2020} dataset including subjects older than 45 years of age only). Notable exceptions include studies training on several publicly available datasets which have reached training sample sizes of up to N=2,001 covering the full adult age range\cite{cole2017a}. Also, more recent studies (ENIGMA\cite{han2020} or Kaufmann et al.\cite{kaufmann2019}) have reached very good performance in large training and validation datasets despite focusing on a limited set of morphological features. Such studies clearly provide major improvements with regard to previous studies.

Second, the assessment of the performance of the trained model is often severely hampered by small validation datasets. With so-called leave-one-out cross validation – a validation scheme which relies on averaging performance across N models using a validation dataset size of N=1 in each iteration – the norm rather than the exception, performance estimates are often highly variable. If independent model validation on previously unseen datasets is conducted at all, validation samples are often small. Since validation sample sizes below N=100 may yield a substantial percentage of spuriously inflated performance estimates, the stability of brain-age estimates underlying many brain-age studies might be questioned\cite{flint2019}. Further, validation on large independent datasets including data from multiple recruitment centers and imaging sites is rare. Notable exceptions include Bashyam et al.\cite{bashyam2020} who used a cross-disorder sample comprising N=11,729 participants. Also, brain-age models are regularly not evaluated regarding potential biases such as gender for which fitting separate models might be beneficial. 

Third, although the initial brain-age framework suggested linear Relevance Vector Regression (RVR), algorithms commonly used today also include (non-linear) Support Vector Machines and, more recently, Gaussian Process Regressors\cite{franke2019}. Often the choice of algorithm is not justified and empirical comparisons of different approaches are rare. Especially problematic is the fact that Gaussian Process Regression and RVR runtimes scale cubically with the number of samples, rendering training on large datasets difficult. Also, models trained using these algorithms cannot be shared amongst researchers due to data protection issues as they allow for a partial or even complete reconstruction of the training data, thereby hampering external validation. Moreover, the lack of publicly available brain-age models forces researchers to use a large portion of their data for brain-age model training and validation – either vastly decreasing statistical power for subsequent analyses or introducing additional variation due to cross-validation. 
Fourth, an analysis regarding which characteristics of the brain drive brain-age predictions is either not conducted at all or results are not comparable across studies due to different preprocessing and algorithm-specific importance score mapping. For example, mapping Support Vector Machine weights as a proxy for feature importance may yield vastly different results than deriving Gaussian Process g-maps\cite{marquand2010}, rendering results incomparable. In addition, studies investigate the properties and performance of the machine learning model, not the interaction with single-subject data underlying predictions, thereby disregarding individual differences driving predictions. With ever growing samples allowing for the application of ever more sophisticated machine learning algorithms, this underscores the need for a principled approach to transparency (cf. Explainability\cite{voosen2017}) that is comparable across algorithms.

\begin{figure}
	\centering
	\includegraphics{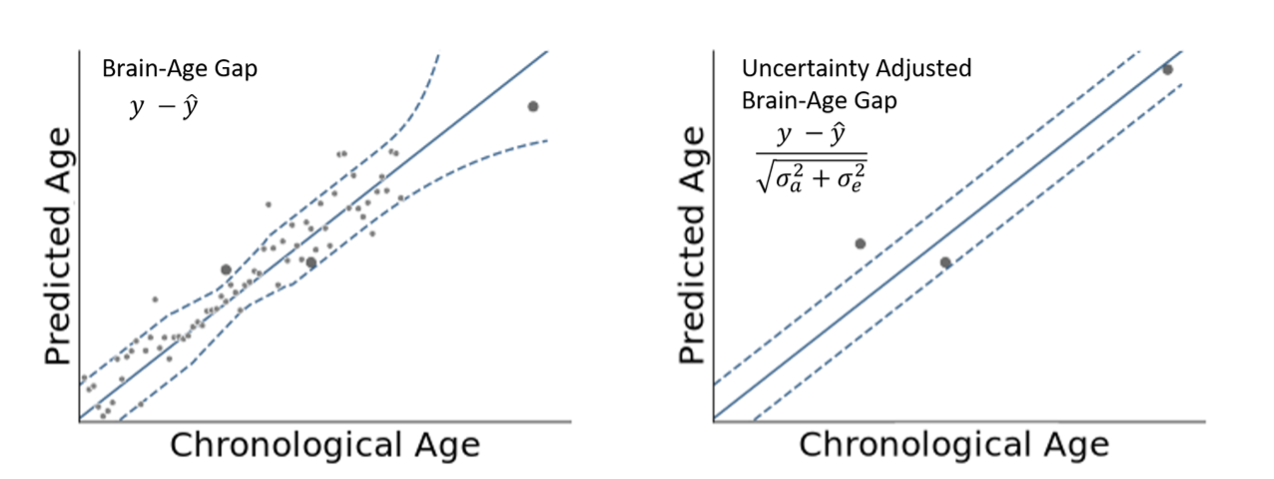}
	\caption{Example data illustrating the effects of adjusting the Brain-Age Gap (BAG) for individual uncertainty. Left panel: Regression model (solid line) with uncertainty estimate (e.g. 95\% predictive interval; dotted lines) trained on toy-data with varying density and variability (light grey) applied to three test samples (dark grey). BAG is defined as a test sample’s distance from the regression line. Right panel: Uncertainty adjustment increases BAG in areas of low uncertainty (left-most test sample) and decreases it in areas of high uncertainty (right-most test sample). $\sigma_a$: aleatory uncertainty, $\sigma_e$: epistemic uncertainty.}
	\label{fig:1}
\end{figure}

Most importantly, the core metric of the field – i.e. the difference between chronological and predicted brain-age (commonly referred to as Brain-Age Gap, BAG) – does not account for uncertainty in model predictions. Not adjusting the BAG for uncertainty, however, renders findings of accelerated or decelerated aging confounded with data density and variability (cf. the concept of Normative Modeling\cite{marquand2016a} for an introduction). Specifically, deviations between chronological age and BAG may arise not only from neural changes as intended, but also erroneously from high uncertainty. Therefore, failing to properly model uncertainty may lead to spurious results, which depend on the characteristics of the training sample and properties of the model rather than on the underlying association of a variable with BAG. Importantly, uncertainty arising from noise inherent in the observations (i.e. aleatory uncertainty) as well as uncertainty arising from the model itself (i.e. epistemic uncertainty) must be considered. Figure \ref{fig:1} illustrates the concept of adjusting the BAG by individual uncertainty.

While not commonly used in brain-age research, algorithms often employed in brain-age modeling such as RVR and the Gaussian Process Regression (GPR) are – in principle – capable of modeling aleatory and epistemic uncertainty. For high-dimensional inputs, however, uncertainty estimation becomes exceedingly difficult using these methods. This has sparked a plethora of research into alternative approaches, especially for neural networks\cite{cannon2018a,gal2016a,palma2020}. While interesting, most of these approaches do not consider aleatory and epistemic uncertainty together and none have been applied to brain-age research.
Here, we address these issues by 1) introducing a robust Monte-Carlo Dropout Composite Quantile Regression (MCCQR) Neural Network architecture capable of estimating aleatory and epistemic uncertainty in high-dimensional data, 2) training our model on anatomical MRI data of N=10,691 individuals between 20 and 72 years of age from the German National Cohort (GNC) , 3) validating the resulting model using leave-site-out cross-validation across ten recruitment centers as well as three additionally independent, external validation sets comprising a total of N=4,004 samples between 18 and 86 years of age, 4) benchmarking the MCCQR model against five most commonly used algorithms in brain-age modeling with regard to predictive performance and quantification of uncertainty, 5) systematically assessing model bias for gender, age, and ethnicity, and 6) developing a unified Explainability approach based on the combination of occlusion-sensitivity mapping and Generalized Linear Multi-Level Modelling to identify brain-regions driving brain-age predictions. Building on data from the GNC study, we apply the MCCQR model to predict uncertainty adjusted brain-age gaps and investigate their association with Body Mass Index (BMI) and Major Depressive Disorder. As training data cannot be reconstructed from the MCCQR model, we make the pre-trained model publicly available for validation and use in future research.

\section{Results}

\underline{Model Performance}\newline
We evaluated our MCCQR Neural Network model against five commonly used algorithms in brain-age modeling – namely the RVR, linear Support Vector Machine (SVM), Support Vector Machine with a Radial Basis Function kernel (SVM-rbf), Gaussian Process Regression (GPR), and Least Absolute Shrinkage and Selection Operator (LASSO) Regression – regarding predictive performance. For comparison, we also evaluated a version of our neural network model without uncertainty quantification, but with an otherwise identical network structure and hyperparameters (Artificial Neural Network, ANN). 

We iteratively trained our model on data of patients recruited by nine of the ten recruitment centers contributing MRI data to the GNC (N=10,691) and predicted brain-age for all samples from the remaining center. This leave-site-out cross-validation showed a Median Absolute Error (MAE) across all ten recruitment centers of 2.94 years (std=.22) for the MCCQR model. Performance of the other algorithms ranged from 3.05 (std=.22) for, both, GPR and SVM to 4.25 (std=.30) for LASSO Regression. Results of 10-fold cross-validation corroborate this ranking of performance with the MCCQR reaching a MAE of 2.95 years (std=.16) with GPR and SVM MAE=3.09 (std=.11) and LASSO Regression MAE=4.19 (std=.11). The ANN obtained MAE=3.10 (std=.14) for leave-site-out cross-validation and MAE=3.02 (std=.15) for 10-fold cross-validation. 

\begin{table}
	\centering
	\begin{tabular}{c c c c c c}
		\hline
		\textbf{Model} & \textbf{Leave-Site-Out CV} & \textbf{10-fold CV} & \textbf{BiDirect} & \textbf{MACS} & \textbf{IXI}\\\hline\hline
		RVR & 3.37 (.16) & 3.32 (.13) & 3.60 & 5.07 & 4.91\\\hline
		GPR & 3.05 (.22) & 3.9 (.11) & 3.74 & 4.15 & 5.03\\\hline
		SVM & 3.05 (.22) & 3.09 (.11) & 3.74 & 4.15 &5.03\\\hline
		SVM-rbf & 4.19 (.27) & 4.16 (.16) & 4.79 & 9.92 & 8.10\\\hline
		LASSO & 4.25 (.30) & 4.19 (.12) & 4.44 & 8.35 & 6.94\\\hline
		ANN & 3.10 (.14) & 3.02 (.15) & 3.56 & \textbf{3.76} & \textbf{4.48}\\\hline
		MCCQR & \textbf{2.94} (.22) & \textbf{2.95} (.16) & \textbf{3.45} & 3.91 & 4.57
	\end{tabular}
	\caption{Median Absolute Error for all models, cross-validation schemes and independent validation samples. (CV: Cross-validation; for cross-validation, standard deviation across folds is given in parentheses)}
	\label{table1}
\end{table}

While cross-validation performance – particularly across recruitment centers – usually provides good estimates of generalization performance, it does not consider additional sources of variability such as different data acquisition protocols, alterations in recruitment or sample characteristics. Therefore, we validated all models in three independent samples (N=4,004), namely the BiDirect study, the Marburg-Münster Affective Disorders Cohort Study (MACS), and the IXI dataset (IXI). To assess stability under real-world conditions of later use, these samples covered a larger age range than the training data (20 to 86 years vs. 20 to 72 years in the GNC sample) as well as less restrictive exclusion criteria (for a detailed description, see Methods): For the BiDirect sample (N=1,460), the MCCQR model reached a MAE of 3.45 years. Performance of the other models ranged from 3.60 for the RVR to 4.79 for the SVM-rbf. The ANN reached a MAE of 3.76 years. In the MACS sample (N=1,986)\cite{vogelbacher2018}, the MCCQR model reached a MAE of 3.92, while the other models obtained generalization performance between 4.15 (GPR and SVM) and 9.92 (SVM-rbf). The ANN reached MAE=3.76 years. Finally, we evaluated performance on the publicly available IXI dataset (www.brain-development.org, N=561). The MCCQR model and the ANN reached a MAE of 4.57 and 4.48 years, respectively. The other models’ performances ranged between MAE=4.91 years (RVR) and MAE=8.10 (SVM-rbf).Table \ref{table1} shows individual model performance for leave-site-out, 10-fold cross-validation and the three independent validation samples. MCCQR achieves lower MAE compared to all other algorithms commonly used in brain-age research. The ANN displays lower MAE than the MCCQR in two out of five cases, indicating a slight advantage for the neural network architecture disregarding uncertainty in these cases.

\begin{figure}
	\centering
	\includegraphics{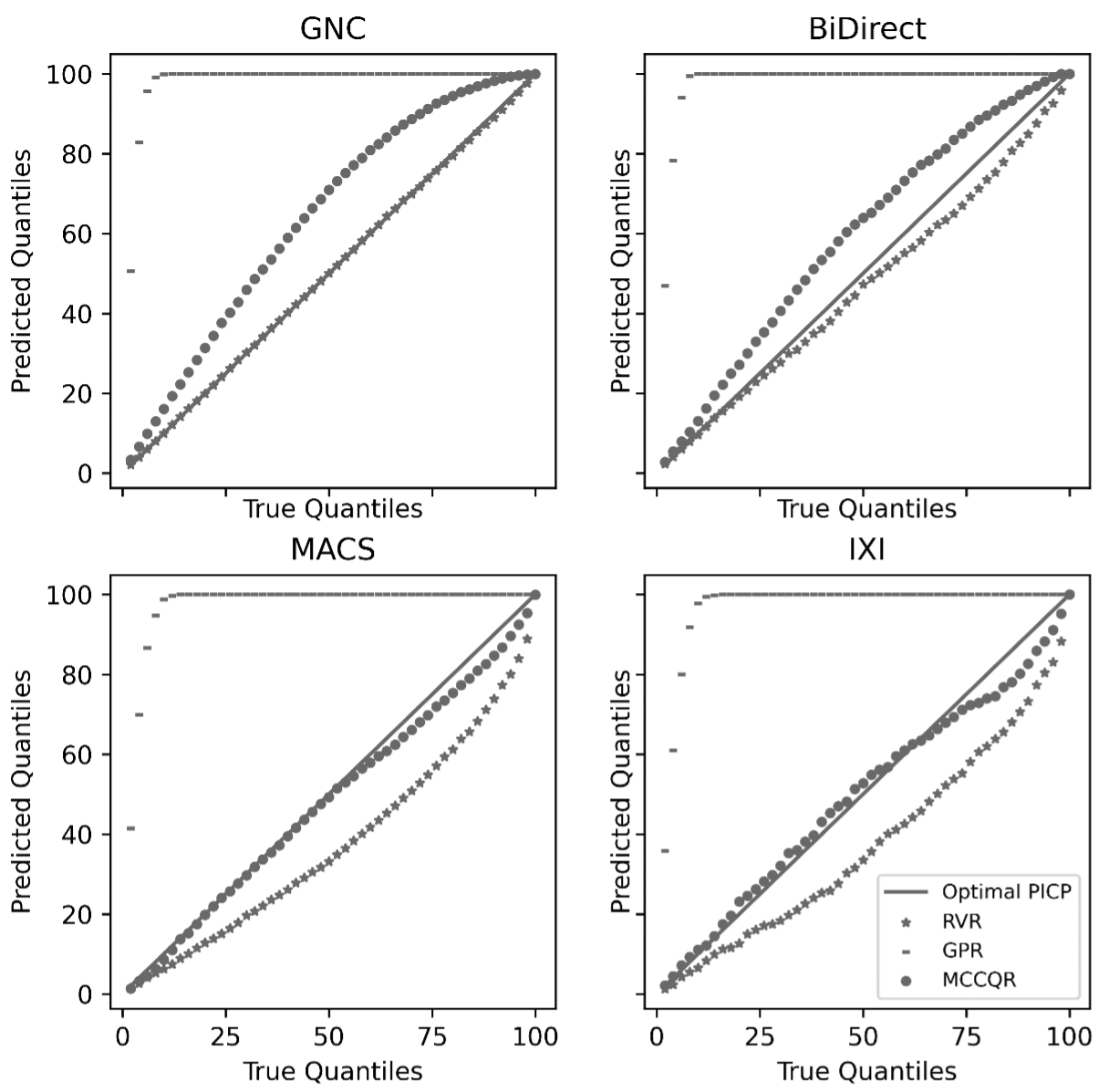}
	\caption{Prediction Interval Coverage Probabilities (PICP) for leave-site-out GNC and independent validation samples (BiDirect, MACS, and IXI) for the Relevance Vector Regression (RVR), the Gaussian Process Regression (GPR), and our Monte Carlo Composite Quantile Regression (MCCQR) neural network. Underestimation (overestimation) of uncertainty occurs, if empirical PICPs are below (above) optimal PICP as indicated by the solid line. }
	\label{fig:2}
\end{figure}

\underline{Uncertainty Quantification}\newline
Adjusting BAG for aleatory and epistemic uncertainty is crucial as findings of accelerated or decelerated aging may otherwise be driven by e.g., training data density and variability. While algorithms such as RVR and GPR are – in principle – capable of modeling aleatory and epistemic uncertainty, performance on high-dimensional data may be problematic. To evaluate the quality of the uncertainty quantification, we estimated uncertainty using the MCCQR, RVR, and GPR (note that the other algorithms do not readily provide uncertainty estimates). Then, we assessed Prediction Interval Coverage Probability (PICP), i.e., the probability that a sample’s true value is contained within the predictive interval. Figure \ref{fig:2} depicts PICPs for given quantile values for cross-validation and the three independent validation datasets. Note that underestimation of uncertainty is highly problematic as samples may be erroneously characterized as deviants from the normal brain-aging trajectory. Overestimation of uncertainty decreases the ability to detect outliers, rendering the approach more conservative. While the GPR substantially overestimates uncertainty in all datasets, RVR and MCCQR provide high-quality uncertainty estimations for GNC and BiDirect datasets. MCCQR outperforms RVR in the MACS and IXI datasets. Note that incorporating epistemic or aleatory uncertainty alone – as has recently been suggested\cite{palma2020} for brain-age models – systematically underestimates uncertainty (see Supplementary Figure S1).

\underline{Relevance of Uncertainty Adjustment}\newline
As outlined above, brain-age gaps may arise not only from brain changes as intended, but also erroneously from high uncertainty. I.e., a person may have a large BAG not only due to actual changes in the brain, but also due to properties of the underlying machine learning model which arise from characteristics of the training data such as data density and variability. Here, we empirically demonstrate two such cases. 

First, we investigated the association of BAG and Body Mass Index (BMI) in the GNC study data. Based on this, we find a significant association between BMI and BAG (F(1,10344)=7.06, p=0.008). However, this effect is no longer observed, if uncertainty is considered by scaling BAG with the standard deviation of the individual predictive distribution (F(1,10344)=0.15, p=0.697). Correspondingly, effect size (partial $\eta^2$) was reduced by 98\%.

Second, we investigated the difference in BAG between a population sample (N=1,612) and patients suffering from Major Depressive Disorder (N= 1,541) from the MACS and the BiDirect sample. While the standard analysis based on BAG failed to reveal a significant difference (F(1,3148)=1.61, p=0.204), the same analysis based on uncertainty corrected BAG detected a significant effect (F(1,3148)=5.59, p=0.018). Likewise, partial $\eta^2$ was increased by 247\%.

\underline{Bias Assessment}\newline
As machine learning models are not programmed, but trained, they will mimic systematic biases inherent in their training data\cite{cearns2019}. While this potential algorithmic bias must be carefully investigated with regard to performance differences in specific subgroups to determine for which populations it yields robust estimates, it is often neglected in brain-age research (for an in-depth discussion, see \cite{fat2018}). Here, we investigated model performance differences for gender, ethnicity, and age.

We found that model performance of the MCCQR did not substantially differ with regard to gender, reaching a standardized MAE of .459 for females and a standardized MAE of .456 for males in the BiDirect dataset (due to different age ranges between males and females, we standardized MAE to make results comparable). In the MACS dataset, standardized MAE reached .298 in females and .264 in males. For the IXI dataset, the model reached a standardized MAE of .264 in females and .300 in males. Note that we defined gender as male and female as no further information on other genders was available. Regarding ethnicity, we tested our model on a publicly available dataset from Beijing Normal University (N=179) between 18 and 28 years of age. Despite the low age range, the MCCQR failed to provide reasonable brain-age predictions with a MAE=8.39. Finally, we investigated model performance across different ages. As commonly reported in brain-age studies, we observed a correlation between BAG and age (r=-.39) as well as between uncertainty corrected BAG and age (r=-.43) in the BiDirect sample. The same is true for the MACS dataset (BAG r=-.36; uncertainty corrected BAG r=-.40) as well as the IXI dataset (BAG r=-.70; uncertainty corrected BAG r=-.75), implying that in all three validation datasets, performance is better in older participants (see Supplementary Figure S2). Thus, we included age as a covariate in all statistical analyses involving the BAG in section Relevance of Uncertainty Adjustment.

\underline{Explainability}\newline
Combining occlusion-sensitivity mapping and Generalized Linear Multi-Level Modelling, we investigated which brain-regions are relevant for accurate brain-age prediction in the MACS sample (N=1,986). We show that occlusion – i.e. the exclusion of data – of any of the 116 regions of the AAL brain-atlas significantly affected model performance (largest p<.001, see Supplementary Table S1). Interestingly, occlusion of 5 regions leads to increased BAG, while occlusion of 107 regions decreased BAG. Strongest BAG-increasing effects were observed for left and right putamen and left Heschl’s gyrus. Strongest BAG-decreasing effects were observed for right inferior temporal gyrus, left middle temporal gyrus and left middle frontal gyrus (see Supplementary Figure S3). 

\section{Discussion}
We trained an uncertainty-aware, shareable and transparent MCCQR Neural Network on N=10,691 samples from the GNC. This model achieves lower error rates compared to existing models across ten recruitment centers and in three additional, independent validation samples (N=4,004). Importantly, the model thus generalized well to independent datasets with larger age range than the training data. In contrast to currently used algorithms – which either do not provide uncertainty estimates or over- or underestimate uncertainty – the MCCQR model provides robust, distribution-free uncertainty quantification in high-dimensional neuroimaging data. Building on this, we show in N=10,691 subjects from the GNC that a spuriously inflated brain-age acceleration effect for Body Mass Index is found if uncertainty is ignored. Likewise, we demonstrate in the BiDirect sample that correcting for uncertainty also increases power to detect accelerated aging in Major Depressive Disorder (N=688) as compared to a population sample (N=719). These findings underscore the need to adjust brain-age gaps for aleatory and epistemic uncertainty. Non-adjustment for these uncertainty components led to false positive as well as false negative findings in our studies. 

While the importance of uncertainty quantification has been discussed at length in the context of normative modeling (cf.\cite {marquand2016a} for a discussion of the relevance of including aleatory and epistemic uncertainty) and two of the algorithms commonly used in brain-age research (mainly GPR) provide uncertainty quantification, it has largely been ignored in the brain-age literature to date. This could be due to several reasons: First, GPR and RVR do not scale well to large datasets as they require the inversion of the kernel matrix and more scalable GPRs based on variational Bayes are still an active area of research. Second, reasonable uncertainty estimation for GPR models becomes exceedingly difficult in situations in which the number of features far exceeds the number of samples as is the case in virtually all MRI studies. This issue also arose in our study, leading to unreasonable uncertainty estimates (cf. Figure 2). Third, GPR and RVR models require the full training sample to make predictions. While scalable GPR approaches only require a subset of the training data, the quality of predictions directly depends on the number and representativity of these so-called induction points. Note that in our study, the RVR algorithm’s uncertainty quantification was of high quality in the GNC and BiDirect data, but decreased in comparison to the MCCQR in the MACS and IXI datasets. This performance might render it a viable alternative with regard to uncertainty quantification in some cases. However, as literally the entire training dataset is required to make predictions using RVR, the data protection issues arising from this largely prohibit model sharing and thus independent validation.

Notably, a recent study also recognized the issue of uncertainty quantification for brain-age modeling and employed quantile regression to estimate aleatory uncertainty in brain-age prediction\cite{palma2020}. While this approach accounts for aleatory uncertainty induced by e.g. measurement error, it does not consider epistemic uncertainty, i.e., uncertainty in the model weights. Empirically, we showed that accounting for aleatory uncertainty only, substantially underestimated true uncertainty (cf. Supplementary Figure 1). If data density differs over age groups as seemingly uncertainty-corrected brain-age gaps may still be confounded. Thereby, deviant brain-ages might spuriously arise from differential training data density – an effect especially problematic given the relatively small training sample sizes employed in most brain-age studies. In addition, it may be difficult to detect as only subsets of BAGs of a given sample are affected.

As mentioned above, our approach is intimately related to normative modeling\cite{marquand2016a}. In contrast to this approach, however, we do not seek to quantify voxel-wise deviation, but deviation on the level of the individual. Hence, we do not predict single voxel data from chronological age, but chronological age from the multivariate pattern of whole-brain data. While this directly yields brain-age predictions on the level of the individual – hence circumventing the need to estimate individual-level predictions based on extreme value statistics or the combination of deviations across voxels as has been suggested for normative modeling – it cannot directly be used as a brain mapping method. To this end, we adopted an occlusion-sensitivity mapping approach, which quantifies regional importance as the reduction in BAG when features from a specific region are withheld\cite{zeiler2014}. Compared to other approaches to Explainability such as the visualization of the network weights using e.g. Layer-wise Relevance Propagation\cite{bach2015}, occlusion-sensitivity mapping comes with the benefit of yielding relevant features for each individual. From a methodological point of view, the multi-level model used in this study holds crucial advantages over the commonly employed mass-univariate approach. First, in the mass-univariate approach, estimates for the effect of the predictors on BAG are only informed by one region. In the multi-level framework, however, each estimate is informed not only by its region but also by all other regions. Second, the multi-level model enables us to include theoretically necessary control variables, which vary over brain regions, but not over samples, such as the size of each region of interest. Third, the multi-level approach controls for dependency within the model, alleviating the need for multiple comparison correction as required in the mass-univariate case\cite{gelman2012}. Interestingly, this analysis revealed that occlusion of any region resulted in a significant change of brain-age predictions, underscoring the multivariate, distributed nature of aging in the brain. Also, our results mitigate concerns that high-performance brain-age models might focus on a small subset of features not affected by pathology\cite{bashyam2020}.
Machine learning models are trained on data and may thus mimic systematic biases inherent in their training data. Investigating performance in specific subgroups revealed that the MCCQR performed comparably for females and males. As is the case for most brain-age models, we observed a correlation between BAG and age. This behavior is not unexpected, as errors in regression modelling will always tend towards the mean (age in this case) of the training set. Nonetheless, it requires modeling age as a covariate in all analyses aiming to associate BAG with variables of interest (for an introduction to age-bias correction approaches in brain-age modeling, see \cite{delange2020}). Investigating ethnicity bias, we showed that our model fails to accurately predict age in a Chinese sample. While this limitation was to be expected – given the GNC aims to recruit a random population sample of Germany was used as the sole training data – it also underscores the need for systematic bias assessment. Importantly, the MCCQR also offers an opportunity to remedy this issue (see below).

The MCCQR – in contrast to most approaches used in brain-age research – does not allow for the reconstruction of individual samples from the training sample. Thus, we make it publicly available. This is beneficial for three reasons: First, it allows others to independently assess MCCQR model performance. Second, it may dramatically increase power and robustness of smaller brain-age studies by circumventing the need to train a brain-age model and test associations with variables of interest in the same dataset. Third, it enables researchers to continue training the model with more data. For example, fine-tune the MCCQR model with data from other ethnic groups – e.g. Asian – would help the model generalize better. In the same vein, our model could be extended by adding e.g., three-dimensional convolution layers (as done e.g. in \cite{cole2017a,bashyam2020}) to the MCCQR to allow predictions directly from raw MRI data without the need for preprocessing. While this study constitutes a first step towards incorporating uncertainty in brain-age modeling, it is limited in several ways. First, we neither evaluated the model on a large sample of older participants (>72 years) nor on a sample of adolescents. Second, we did not explicitly model GNC imaging sites during training and a benchmark against state-of-the-art Deep Learning approaches is missing. Future studies could therefore not only evaluate the model, but also increase generalization and usability by training on more diverse datasets and developing model architecture. We facilitate this research by making the pre-trained model publicly available with this publication.

In light of this, we provide the uncertainty-aware, shareable, and transparent MCCQR architecture and pre-trained model with the intention stimulate further research and increase power for small-sample analyses. The pre-trained PHOTON-AI model and code can be downloaded from the PHOTON AI model repository (\href{www.photon-ai.com/repo}{www.photon-ai.com/repo}).

\section{Materials and Methods}
\underline{Training and Validation Samples}\newline
Whole-brain MRI data from five sources were used. We trained the MCCQR on the German National Cohort (GNC) sample. Results for leave-site-out and 10-fold cross-validation are also based upon the GNC sample. Independent validation was based upon the BiDirect sample, the MACS data, and the IXI dataset. Ethnicity bias assessment was conducted using the Beijing Normal University dataset (see below). In the following, we describe each dataset in more details. Also, SupplementaryTable S2 provides further sample characteristics, including sample sizes, gender distribution, age minimum and maximum as well as standard deviation.

\underline{German National Cohort (GNC):} This cohort is one of the population based 'megacohorts' and examined 205,000 Germans, aged 20 to 72 years in 18 study centers across Germany between 2014 and 2019 using a comprehensive program. Specifically, this included a 3.0 Tesla whole-body MRI (T1w-MPRAGE) in 30.000 participants, performed in five GNC imaging centers equipped with dedicated identical magnets (Skyra, Siemens Healthineers, Erlangen, Germany) examining participants from 11 of the centers. This analysis is based on the 'data freeze 100K' milestone for the first 100,000 participants which also included the first 10,691 participants with completed MRIs of sufficient quality (for a detailed protocol, see \cite{consortium2014,bamberg2015}). We calculated body mass index from directly measured height and weight (kg/m2; mean BDI = 26.82; std BDI = 4.76). To ensure that our models were not driven by data quality or total intracranial volume, we assessed the predictive power of the three data quality parameters provided by the Cat12 toolbox as well as Total Intracranial Volume (TIV). We show that TIV, Bias, Noise, and Weighted Average IQR combined explain only 9.06\% of variation in age using a linear Support Vector Machine with 10-fold cross-validation compared to the 86\% achieved by the MCCQR model.

\underline{BiDirect:} The BiDirect study is an ongoing study that comprises three distinct cohorts: patients hospitalized for an acute episode of major depression, patients two to four months after an acute cardiac event and healthy controls randomly drawn from the population register of the city of Münster, Germany. Baseline examination of all participants included a structural MRI of the brain, a computer-assisted face-to-face interview about sociodemographic characteristics, a medical history, an extensive psychiatric assessment and collection of blood samples. Inclusion criteria for the present study were availability of completed baseline MRI data with sufficient MRI quality. All patients with Major Depressive Disorder had an episode of major depression at the time of recruitment and were either currently hospitalized (> 90\%) or had been hospitalized for depression at least once during the twelve months before inclusion in the study (< 10\%). Further details on the rationale, design and recruitment procedures of the BiDirect study have been described elsewhere\cite{teismann2014}.

\underline{Marburg-Münster Affective Disorders Cohort Study (MACS):} Participants were recruited through psychiatric hospitals or newspaper advertisements. Inclusion criteria included mild, moderate or partially remitted Major Depressive Disorder episodes in addition to severe depression. Patients could be undergoing inpatient, outpatient or no current treatment. The MACS was conducted at two imaging sites – University of Münster, Germany and University of Marburg, Germany. Further details about the structure of the MACS\cite{kircher2019} and MRI quality assurance protocol\cite{vogelbacher2018} are provided elsewhere. Inclusion criteria for the present study were availability of completed baseline MRI data with sufficient MRI quality (see \cite{vogelbacher2018} for details).

\underline{Information eXtraction from Images (IXI):} This dataset comprises images from normal, healthy subjects, along with demographic characteristics, collected as part of the Information eXtraction from Images (IXI) project available for download (https://brain-development.org/ixi-dataset/). The data has been collected at three hospitals in London (Hammersmith Hospital using a Philips 3T system), Guy’s Hospital using a Philips 1.5T system, and Institute of Psychiatry using a GE 1.5T system). Inclusion criteria for the present study were availability of completed baseline MRI data.

\underline{Beijing Normal University:} This dataset includes 180 healthy controls from a community (student) sample at Beijing Normal University in China. Inclusion criteria for the present study were availability of completed baseline MRI data. Further details can be found online (\href{http://fcon_1000.projects.nitrc.org/indi/retro/BeijingEnhanced.html}{http://fcon\_1000.projects.nitrc.org/indi/retro/BeijingEnhanced.html}).

\underline{Magnetic Resonance Imaging (MRI) Preprocessing}\newline
MRI data was preprocessed using the CAT12 toolbox (built 1450 with SPM12 version 7487 and Matlab 2019a; http://dbm.neuro.uni-jena.de/cat) with default parameters. Images were bias-corrected, segmented using tissue classification, normalized to MNI-space using DARTEL-normalization, smoothed with an isotropic Gaussian Kernel (8mm FWHM), and resampled to 3mm isomorphic voxels. Using the PHOTON AI software (see section Model Training, Validation, and Mapping below), a whole-brain mask comprising all grey-matter voxels was applied, data was vectorized, features with zero-variance in the GNC dataset were removed, and the scikit-learn Standard Scaler was applied.

\underline{Monte Carlo Dropout Composite Quantile Regression (MCCQR Model)}\newline
Commonly, regression models $f^W(X)$ are used to describe the relationship of features $X=(x_1,\dots,x_n)$ and a target variable $Y=(y_1,\dots,y_n)$. We denote model predictions $\hat{y}=f^W(x)$ with $W$ the model parameters. Commonly, such regression models provide predictions as point estimates rather predictive distributions. We thus consider two types of uncertainty in accordance with Kendall and Gal\cite{kendall2017}. The first – aleatory uncertainty – captures noise inherent in the observations. The second, epistemic uncertainty, accounts for uncertainty in the model. While the former is irreducible for a given model, the latter depends on data availability and training and is thus affected by data density and can be reduced by training 1) the model based on the loss function and 2) with additional data. While numerous approaches to uncertainty quantification have been suggested, no commonly accepted approach exists. While Bayesian approaches such as Gaussian Process Regression provide a mathematically elegant framework, such approaches often fail in high-dimensional settings or are not scalable to large datasets\cite{liu2020}. In particular, capturing aleatory and epistemic uncertainty within the same neural network model remains challenging\cite{tagasovska2018a}. Here, we suggest to combine Composite Quantile Regression and Monte Carlo Dropout to model aleatory and epistemic uncertainty within a single framework, respectively. 

\underline{Composite Quantile Regression:} Quantile regression (QR) provides an estimate not only of the conditional mean (as is done when optimizing for Mean Absolute Error), but of any quantile in the data. This comes with two advantages. First, we can estimate the median (instead of the mean), thereby obtaining a prediction more robust to potential outliers. Second, predicting quantiles can yield predictive intervals, thereby modeling aleatory uncertainty. For example, we cannot only predict a sample’s target value, but also the 95\% confidence bounds of this prediction. This makes QR interesting whenever percentile curves are of interest, e.g. when screening for abnormal growth. 

Commonly, conditional quantiles for pre-determined quantile probabilities are estimated separately by different regression equations. These are then combined to build a piecewise estimate of the conditional response distribution. As this approach is prone to “quantile crossing”, i.e. QR predictions do not increase with the specified quantile probability $\tau$, Composite QR was introduced\cite{xu2017}. In Composite QR, simultaneous estimates for multiple values of $\tau$ are obtained and the regression coefficients are shared across the different quantile regression models. In essence, we aim to approximate a single $\tau$-independent function that best describes the function to be learned. Structurally, Composite QR differs from QR only in that the QR error (tilted loss) function is summed over many, usually equally spaced values of $\tau$. Specifically, QR is implemented using the quantile regression error function\cite{koenker1978} to optimize a neural network via
\begin{equation}
	E_\tau = \frac{1}{N}\sum_{t=1}^{N}\mathfrak{p}_\tau(y(t) - \hat{y}(t))
\end{equation}
with the tilted loss function $\mathfrak{p}$
\begin{equation}
	\mathfrak{p}_\tau=\tau\cdot\epsilon \text{ if } \epsilon\ge0, \text{ else } (1-\tau)\epsilon
\end{equation}
Composite QR extends this idea to estimating multiple quantiles simultaneously. This is not only more stable, but also computationally more efficient as fewer coefficients and fewer operations during weight updating are required. To achieve this, we modified the QR error function (above) to
\begin{equation}
	E_{C\tau}=\frac{1}{KN}\sum_{k=1}^{K}\sum_{t=1}^{N}\mathfrak{p}_{\tau_k}(y(t)-\hat{y}(t))
\end{equation}
where $\tau_k$  are usually equally spaced, for example $\tau_k= \frac{k}{K+1}$  for $k=1, 2,\dots, K$  as suggested by Cannon\cite{cannon2018a}. Specifically, we calculate 101 equally spaced quantile values for  . To allow for continuous sampling during prediction, we linearly interpolated these quantile values. Note that while this approach does not formally guarantee the absence of quantile cross-over, Composite QR reduced the likelihood so much, that we never empirically observed quantile cross-over for any of the more than 1.48 billion quantiles (arising from 101 quantiles, 14,695 samples, and 1,000 draws from the predictive distribution per sample) estimated during independent and cross-validation in this study.

\underline{Monte Calo Dropout:} While Composite QR captures aleatory uncertainty, it does not account for epistemic uncertainty, i.e. uncertainty in the model parameters. For example, epistemic uncertainty should be higher in regions of the input space where little data is available whereas it should be lower in regions with high data density. While we could in principle model each weight of a Neural Network as distribution from which to sample weight values during prediction, estimating such probabilistic models remains challenging for high-dimensional data. Following Gal and Ghahramani\cite{gal2016a}, we therefore adopted a Monte Carlo Dropout approach. While dropout is commonly employed for regularization during Neural Network training, the Monte Carlo Dropout approach enforces dropout during training and at test time. Thus, dropout can be used to obtain   predictions for the same input with different active neurons. This allows the estimation of $p(y|f^W(x))$, the mean probability of a prediction given a test input $X=(x_1,\dots,x_n)$ for the neural network $f$ with the according weights $W$. We therefore define our likelihood as a Gaussian with mean given by the model output according to Gal and Ghahramani\cite{gal2016a}:
\begin{equation}
	p(y|f^W(x)) = N(f^W(x), \sigma^2)
\end{equation}
We can then calculate the mean probability using Monto Carlo Dropout performing $T$ forward passes using randomly sampled weight values $W_i$ from the neural network using dropout as
\begin{equation}
	p(y|f^W(x))=\frac{1}{T}\sum_{i=1}^{T}p(f^W(x)=k|x, W_i)
\end{equation}

\underline{Monte Carlo Dropout Composite Quantile Regression (MCCQR):} To model epistemic and aleatory uncertainty simultaneously within the same framework, we combined Monte Carlo Dropout and Composite Quantile Regression described above. Specifically, we implemented the resulting Monte-Carlo Dropout Composite Quantile Regression Neural Network consisting of one hidden layer with 32 Rectified Linear Units (ReLUs) using Tensorflow 2.0 together with Tensor Flow Probability for robust median MAE calculation over batches. Note that we used Median Absolute Error instead of the more common Mean Absolute Error as the median is more robust to outliers compared to the mean. We trained for 10 epochs with a learning rate of .01, a batch size of 64, and a dropout rate of .2 using the Adam Optimizer with default settings. Predictions were obtained by sampling 1,000 times from each sample’s predictive distribution with random $\tau$ values and dropout enabled. A sample’s brain-age prediction was computed as the median of the resulting values. Likewise, uncertainty was computed as the standard deviation of the resulting values. 

\underline{Reconstruction of individual-level data:} Machine Learning models traditionally used in brain-age modeling allow for the reconstruction of individual-level data. For our model, however, only saving model parameters and network architecture is required. Reconstructing individual-level data from this information alone is not possible for quantitative as well as conceptual reasons: Considering the quantity of information, our final model contains 1,269,701 parameters, 64 of which are not trainable. The complete training dataset consist of 10,696 images containing 39,904 Voxels each, resulting in 426,733,376 parameters in the dataset. This amounts to about 336 times the number of parameters in our model, rendering it incapable of “memorizing” the data. Considering the conceptual nature of the training process, our model applies Monte Carlo Dropout. This method – used to counter overfitting and to estimate epistemic uncertainty – randomly reduces the number of active units during a forward pass. As the network parameters are optimized for inference across the whole training set, the network is not capable of compressing the images. Thus, we cannot recover individual trainings samples from the network.

\underline{Benchmarking Alternative Machine Learning Models}\newline
We evaluated our MCCQR Neural Network model against five commonly used algorithms in brain-age modeling – namely the RVR, linear Support Vector Machine (SVM), Support Vector Machine with a Radial Basis Function kernel (SVM-rbf), Gaussian Process Regression (GPR), and Least Absolute Shrinkage and Selection Operator (LASSO) Regression. We used the fast-rvm implementation from sklearn-bayes (https://github.com/AmazaspShumik/sklearn-bayes) and employed SVM, GPR, and LASSO implementations from sci-kit learn\cite{pedregosa2011} with default settings. For comparison, we also evaluated a version of our neural network model without uncertainty quantification, but with an otherwise identical network structure and hyperparameters optimizing Mean Absolute Error over predictions instead of the tilted loss function used for the MCCQR model.

\underline{Model Training and Validation}\newline
All models were trained and cross-validated using the PHOTON AI software (\href{www.photon-ai.com}{www.photon-ai.com}) for leave-site-out and 10-fold cross-validation on the GNC sample. Independent validation was conducted using PHOTON AI’s .photon format for pipeline-based prediction with the BiDirect, MACS, IXI, and Beijing Normal University samples. 

\underline{Generalized Linear Multi-Level Modelling for Occlusion-Sensitivity Mapping}\newline
Occlusion-Sensitivity Mapping – in analogy to occlusion-based approaches for 2d images\cite{zeiler2014} – quantifies regional importance as the reduction in performance when features from a specific region are withheld. Occlusion was implemented by sequentially setting all voxels within each of the 116 regions of interest (ROIs) of the AAL brain-atlas\cite{tzourio-mazoyer2002} to zero. Occlusion sensitivity is widely used to gain insight into which regions of an image a machine learning model utilizes for prediction. A region is considered more important if model performance decreases more strongly when information from this region is withheld. 

To quantify this notion of importance, we combined Occlusion-Sensitivity Mapping with Generalized Linear Multi-Level Modelling. Specifically, we employed Multiple Linear Regression using the R packages lme4 to model the difference between BAG based on whole-brain data and BAG if information from a specific atlas region is withheld. 
To investigate regionally specific effects, we computed uncertainty corrected BAG estimates for each individual and region of interest (i.e. 116 regions from the AAL atlas distributed with SPM; \href{https://www.fil.ion.ucl.ac.uk/spm/}{https://www.fil.ion.ucl.ac.uk/spm/}) independently. Then, we predicted uncertainty corrected BAG from a group factor with AAL regions as factor levels, controlling for chronological age, site (where appropriate), gender as well as ROI-size. The ROI group factor was defined as treatment contrast with the whole brain (no occlusion) level as reference factor. This way, ROI effect estimates can easily be interpreted as difference in BAG from the full model. P-values for ROI factor level effects were computed using the Kenward-Roger approximation\cite{halekoh2014} as implemented in the afex R package. This approach allows us to test whether the occlusion of a given region results in an above-chance change of predictive performance.

While beyond the scope of this paper, the combination of occlusion-sensitivity mapping and Generalized Linear Multi-Level Modelling also allows for the investigation of regional effects between clinical groups.

More generally, our approach can be considered a form of model-independent Explainable AI: In recent years, a large number of methods have been proposed that aim to shed light on the contribution of variables or groups of variables on a models prediction (for a review and best practice, see\cite{Samek.2021}). Among those are algorithm-specific approaches such as Layer-wise Relevance Propagation which work with Deep Neural Networks or the weight-mapping for Support Vector Machines. In essence, such techniques provide (qualitative) heat maps; i.e. help to understand the informational flow in the network. In contrast, more general approaches ask how the prediction for a single sample/participant would change if the data changed (e.g. if we did not have variable x or if its single-to-noise ratio changes). As we aimed to quantify the effect of omitting/occluding regions of interest independent of a specific training sample, we opted for occlusion mapping here.

\bibliographystyle{unsrt}  
\newpage
\bibliography{MCCQRNN} 

\begin{thebibliography}{10}

\bibitem{cole2017}
James~H. Cole and Katja Franke.
\newblock {Predicting Age Using Neuroimaging: Innovative Brain Ageing
  Biomarkers}.
\newblock {\em Trends in Neurosciences}, 40(12):681---690, 2017.

\bibitem{ludwig1980}
Frederic~C. Ludwig and Mary~E. Smoke.
\newblock {The measurement of biological age}.
\newblock {\em Experimental Aging Research}, 6(6):497---522, 1980.

\bibitem{franke2019}
Katja Franke and Christian Gaser.
\newblock {Ten years of brainage as a neuroimaging biomarker of brain aging:
  What insights have we gained?}
\newblock {\em Frontiers in Neurology}, 10(JUL):789, 2019.

\bibitem{cole2018}
J.~H. Cole, S.~J. Ritchie, M.~E. Bastin, M.~C.~Valdes Hernandez, S.~Munoz
  Maniega, N.~Royle, J.~Corley, A.~Pattie, S.~E. Harris, Q.~Zhang, N.~R. Wray,
  P.~Redmond, R.~E. Marioni, J.~M. Starr, S.~R. Cox, J.~M. Wardlaw, D.~J.
  Sharp, and I.~J. Deary.
\newblock {Brain age predicts mortality}.
\newblock {\em Molecular Psychiatry}, 23(5):1385---1392, 2018.

\bibitem{Bittner.2021}
Nora Bittner, Christiane Jockwitz, Katja Franke, Christian Gaser, Susanne
  Moebus, Ute~J. Bayen, Katrin Amunts, and Svenja Caspers.
\newblock {When your brain looks older than expected: combined lifestyle risk
  and BrainAGE}.
\newblock {\em Brain Structure and Function}, pages 1--25, 2021.

\bibitem{franke2012}
Katja Franke and Christian Gaser.
\newblock {Longitudinal changes in individual BrainAGE in healthy aging, mild
  cognitive impairment, and Alzheimer's Disease}.
\newblock {\em GeroPsych: The Journal of Gerontopsychology and Geriatric
  Psychiatry}, 25(4):235---245, 2012.

\bibitem{gaser2013}
Christian Gaser, Katja Franke, Stefan Kloppel, Nikolaos Koutsouleris, and
  Heinrich Sauer.
\newblock {BrainAGE in Mild Cognitive Impaired Patients: Predicting the
  Conversion to Alzheimer's Disease}.
\newblock {\em PLoS ONE}, 8(6):e67346, 2013.

\bibitem{cole2019}
James~H. Cole, Riccardo~E. Marioni, Sarah~E. Harris, and Ian~J. Deary.
\newblock {Brain age and other bodily ‘ages': implications for
  neuropsychiatry}.
\newblock {\em Molecular Psychiatry}, 24(2):266---281, 2019.

\bibitem{cole2020}
James~H. Cole.
\newblock {Multimodality neuroimaging brain-age in UK biobank: relationship to
  biomedical, lifestyle, and cognitive factors}.
\newblock {\em Neurobiology of Aging}, 92:34---42, 2020.

\bibitem{cole2017a}
James~H. Cole, Rudra~P.K. Poudel, Dimosthenis Tsagkrasoulis, Matthan~W.A. Caan,
  Claire Steves, Tim~D. Spector, and Giovanni Montana.
\newblock {Predicting brain age with deep learning from raw imaging data
  results in a reliable and heritable biomarker}.
\newblock {\em NeuroImage}, 163:115---124, 2017.

\bibitem{han2020}
Laura~K.M. Han, Richard Dinga, Tim Hahn, Christopher~R.K. Ching, Lisa~T. Eyler,
  Lyubomir Aftanas, Moji Aghajani, Andre Aleman, Bernhard~T. Baune, Klaus
  Berger, Ivan Brak, Geraldo~Busatto Filho, Angela Carballedo, Colm~G.
  Connolly, Baptiste Couvy-Duchesne, Kathryn~R. Cullen, Udo Dannlowski,
  Christopher~G. Davey, Danai Dima, Fabio~L.S. Duran, Verena Enneking, Elena
  Filimonova, Stefan Frenzel, Thomas Frodl, Cynthia~H.Y. Fu, Beata~R.
  Godlewska, Ian~H. Gotlib, Hans~J. Grabe, Nynke~A. Groenewold, Dominik
  Grotegerd, Oliver Gruber, Geoffrey~B. Hall, Ben~J. Harrison, Sean~N. Hatton,
  Marco Hermesdorf, Ian~B. Hickie, Tiffany~C. Ho, Norbert Hosten, Andreas
  Jansen, Claas Kahler, Tilo Kircher, Bonnie Klimes-Dougan, Bernd Kramer, Axel
  Krug, Jim Lagopoulos, Ramona Leenings, Frank~P. MacMaster, Glenda MacQueen,
  Andrew McIntosh, Quinn McLellan, Katie~L. McMahon, Sarah~E. Medland, Bryon~A.
  Mueller, Benson Mwangi, Evgeny Osipov, Maria~J. Portella, Elena Pozzi,
  Liesbeth Reneman, Jonathan Repple, Pedro~G.P. Rosa, Matthew~D. Sacchet,
  Philipp~G. Samann, Knut Schnell, Anouk Schrantee, Egle Simulionyte, Jair~C.
  Soares, Jens Sommer, Dan~J. Stein, Olaf Steinstrater, Lachlan~T. Strike,
  Sophia~I. Thomopoulos, Marie Jose~van Tol, Ilya~M. Veer, Robert~R.J.M.
  Vermeiren, Henrik Walter, Nic J.A. van~der Wee, Steven J.A. van~der Werff,
  Heather Whalley, Nils~R. Winter, Katharina Wittfeld, Margaret~J. Wright,
  Mon~Ju Wu, Henry Volzke, Tony~T. Yang, Vasileios Zannias, Greig I.~de
  Zubicaray, Giovana~B. Zunta-Soares, Christoph Abe, Martin Alda, Ole~A.
  Andreassen, Erlend Ben, Caterina~M. Bonnin, Erick~J. Canales-Rodriguez, Dara
  Cannon, Xavier Caseras, Tiffany~M. Chaim-Avancini, Torbjrn Elvsaashagen,
  Pauline Favre, Sonya~F. Foley, Janice~M. Fullerton, Jose~M. Goikolea,
  Bartholomeus~C.M. Haarman, Tomas Hajek, Chantal Henry, Josselin Houenou,
  Fleur~M. Howells, Martin Ingvar, Rayus Kuplicki, Beny Lafer, Mikael Landen,
  Rodrigo Machado-Vieira, Ulrik~F. Malt, Colm McDonald, Philip~B. Mitchell,
  Leila Nabulsi, Maria Concepcion~Garcia Otaduy, Bronwyn~J. Overs, Mircea
  Polosan, Edith Pomarol-Clotet, Joaquim Radua, Maria~M. Rive, Gloria Roberts,
  Henricus~G. Ruhe, Raymond Salvador, Salvador Sarro, Theodore~D.
  Satterthwaite, Jonathan Savitz, Aart~H. Schene, Peter~R. Schofield,
  Mauricio~H. Serpa, Kang Sim, Marcio~Gerhardt Soeiro-de Souza, Ashley~N.
  Sutherland, Henk~S. Temmingh, Garrett~M. Timmons, Anne Uhlmann, Eduard Vieta,
  Daniel~H. Wolf, Marcus~V. Zanetti, Neda Jahanshad, Paul~M. Thompson, Dick~J.
  Veltman, Brenda~W.J.H. Penninx, Andre~F. Marquand, James~H. Cole, and Lianne
  Schmaal.
\newblock {Brain aging in major depressive disorder: results from the ENIGMA
  major depressive disorder working group}.
\newblock {\em Molecular Psychiatry}, pages 1--16, 2020.

\bibitem{kaufmann2019}
Tobias Kaufmann, Dennis van~der Meer, Nhat~Trung Doan, Emanuel Schwarz,
  Martina~J. Lund, Ingrid Agartz, Dag Alns, Deanna~M. Barch, Ramona
  Baur-Streubel, Alessandro Bertolino, Francesco Bettella, Mona~K. Beyer,
  Erlend Ben, Stefan Borgwardt, Christine~L. Brandt, Jan Buitelaar,
  Elisabeth~G. Celius, Simon Cervenka, Annette Conzelmann, Aldo
  Cordova-Palomera, Anders~M. Dale, Dominique J.F.~de Quervain, Pasquale~Di
  Carlo, Srdjan Djurovic, Erlend~S. Drum, Sarah Eisenacher, Torbjrn
  Elvsaashagen, Thomas Espeseth, Helena Fatouros-Bergman, Lena Flyckt, Barbara
  Franke, Oleksandr Frei, Beathe Haatveit, Asta~K. Haaberg, Hanne~F. Harbo,
  Catharina~A. Hartman, Dirk Heslenfeld, Pieter~J. Hoekstra, Einar~A. Hgestl,
  Terry~L. Jernigan, Rune Jonassen, Erik~G. Jonsson, Lars Farde, Lena Flyckt,
  Goran Engberg, Sophie Erhardt, Helena Fatouros-Bergman, Simon Cervenka, Lilly
  Schwieler, Fredrik Piehl, Ingrid Agartz, Karin Collste, Pauliina Victorsson,
  Anna Malmqvist, Mikael Hedberg, Funda Orhan, Peter Kirsch, Iwona Koszewska,
  Knut~K. Kolskaar, Nils~Inge Landr, Stephanie~Le Hellard, Klaus~Peter Lesch,
  Simon Lovestone, Arvid Lundervold, Astri~J. Lundervold, Luigi~A. Maglanoc,
  Ulrik~F. Malt, Patrizia Mecocci, Ingrid Melle, Andreas Meyer-Lindenberg,
  Torgeir Moberget, Linn~B. Norbom, Jan~Egil Nordvik, Lars Nyberg, Jaap
  Oosterlaan, Marco Papalino, Andreas Papassotiropoulos, Paul Pauli, Giulio
  Pergola, Karin Persson, Genevieve Richard, Jaroslav Rokicki, Anne~Marthe
  Sanders, Geir Selbk, Alexey~A. Shadrin, Olav~B. Smeland, Hilkka Soininen,
  Piotr Sowa, Vidar~M. Steen, Magda Tsolaki, Kristine~M. Ulrichsen, Bruno
  Vellas, Lei Wang, Eric Westman, Georg~C. Ziegler, Mathias Zink, Ole~A.
  Andreassen, and Lars~T. Westlye.
\newblock {Common brain disorders are associated with heritable patterns of
  apparent aging of the brain}.
\newblock {\em Nature Neuroscience}, 22(10):1617---1623, 2019.

\bibitem{flint2019}
Claas Flint, Micah Cearns, Nils Opel, Ronny Redlich, David M.~A. Mehler, Daniel
  Emden, Nils~R. Winter, Ramona Leenings, Simon~B. Eickhoff, Tilo Kircher, Axel
  Krug, Igor Nenadic, Volker Arolt, Scott Clark, Bernhard~T. Baune, Xiaoyi
  Jiang, Udo Dannlowski, and Tim Hahn.
\newblock {Systematic Overestimation of Machine Learning Performance in
  Neuroimaging Studies of Depression}.
\newblock 2019.

\bibitem{bashyam2020}
Vishnu~M Bashyam, Guray Erus, Jimit Doshi, Mohamad Habes, Ilya Nasralah, Monica
  Truelove-Hill, Dhivya Srinivasan, Liz Mamourian, Raymond Pomponio, Yong Fan,
  Lenore~J Launer, Colin~L Masters, Paul Maruff, Chuanjun Zhuo, Henry Volzke,
  Sterling~C Johnson, Jurgen Fripp, Nikolaos Koutsouleris, Theodore~D
  Satterthwaite, Daniel Wolf, Raquel~E Gur, Ruben~C Gur, John Morris, Marilyn~S
  Albert, Hans~J Grabe, Susan Resnick, R~Nick Bryan, David~A Wolk, Haochang
  Shou, and Christos Davatzikos.
\newblock {MRI signatures of brain age and disease over the lifespan based on a
  deep brain network and 14 468 individuals worldwide}.
\newblock {\em Brain}, 143(7):1---13, 2020.

\bibitem{marquand2010}
Andre Marquand, Matthew Howard, Michael Brammer, Carlton Chu, Steven Coen, and
  Janaina Mourao-Miranda.
\newblock {Quantitative prediction of subjective pain intensity from
  whole-brain fMRI data using Gaussian processes}.
\newblock {\em NeuroImage}, 49(3):2178---2189, 2010.

\bibitem{voosen2017}
P~Voosen.
\newblock {How AI detectives are craking open the black box of deep learning}.
\newblock {\em Science}, 2017.

\bibitem{marquand2016a}
Andre~F. Marquand, Iead Rezek, Jan Buitelaar, and Christian~F. Beckmann.
\newblock {Understanding Heterogeneity in Clinical Cohorts Using Normative
  Models: Beyond Case-Control Studies}.
\newblock {\em Biological Psychiatry}, 80(7):552---561, 2016.

\bibitem{cannon2018a}
Alex~J. Cannon.
\newblock {Non-crossing nonlinear regression quantiles by monotone composite
  quantile regression neural network, with application to rainfall extremes}.
\newblock {\em Stochastic Environmental Research and Risk Assessment},
  32(11):3207---3225, 2018.

\bibitem{gal2016a}
Yarin Gal and Zoubin Ghahramani.
\newblock {Dropout as a Bayesian Approximation: Representing Model Uncertainty
  in Deep Learning}.
\newblock In Kilian~Q. Weinberger and Maria~Florina Balcan, editors, {\em
  Proceedings of The 33rd International Conference on Machine Learning},
  volume~48, pages 1050 --- 1059. PMLR, 2016.

\bibitem{palma2020}
Marco Palma, Shahin Tavakoli, Julia Brettschneider, and Thomas~E. Nichols.
\newblock {Quantifying uncertainty in brain-predicted age using scalar-on-image
  quantile regression}.
\newblock {\em NeuroImage}, 219:116938, 2020.

\bibitem{vogelbacher2018}
Christoph Vogelbacher, Thomas~W.D. Mobius, Jens Sommer, Verena Schuster, Udo
  Dannlowski, Tilo Kircher, Astrid Dempfle, Andreas Jansen, and Miriam~H.A.
  Bopp.
\newblock {The Marburg-Munster Affective Disorders Cohort Study (MACS): A
  quality assurance protocol for MR neuroimaging data}.
\newblock {\em NeuroImage}, 172(December 2017):450---460, 2018.

\bibitem{cearns2019}
Micah Cearns, Tim Hahn, and Bernhard~T. Baune.
\newblock {Recommendations and future directions for supervised machine
  learning in psychiatry}.
\newblock {\em Translational Psychiatry}, 9(1):271, 2019.

\bibitem{fat2018}
M~L FAT.
\newblock {Fairness, Accountability, and Transparency in Machine Learning}.
\newblock {\em Retrieved December}, 24:2018, 2018.

\bibitem{zeiler2014}
Matthew~D. Zeiler and Rob Fergus.
\newblock {Visualizing and understanding convolutional networks}.
\newblock In {\em Lecture Notes in Computer Science (including subseries
  Lecture Notes in Artificial Intelligence and Lecture Notes in
  Bioinformatics)}, volume 8689 LNCS of {\em Lecture Notes in Computer
  Science}, pages 818---833. 2014.

\bibitem{bach2015}
Sebastian Bach, Alexander Binder, Gregoire Montavon, Frederick Klauschen,
  Klaus~Robert Muller, and Wojciech Samek.
\newblock {On pixel-wise explanations for non-linear classifier decisions by
  layer-wise relevance propagation}.
\newblock {\em PLoS ONE}, 10(7):1---46, 2015.

\bibitem{gelman2012}
Andrew Gelman, Jennifer Hill, and Masanao Yajima.
\newblock {Why We (Usually) Don't Have to Worry About Multiple Comparisons}.
\newblock {\em Journal of Research on Educational Effectiveness},
  5(2):189---211, 2012.

\bibitem{delange2020}
Ann Marie G.~de Lange and James~H. Cole.
\newblock {Commentary: Correction procedures in brain-age prediction}.
\newblock {\em NeuroImage: Clinical}, 26(February):24---26, 2020.

\bibitem{consortium2014}
GNC Consortium; German National Cohort~(GNC) Consortium.
\newblock {The German National Cohort: Aims, study des}.
\newblock {\em European Journal of Epidemiology}, 29(5):371---382, 2014.

\bibitem{bamberg2015}
Fabian Bamberg, Hans-Ulrich Kauczor, Sabine Weckbach, Christopher~L Schlett,
  Michael Forsting, Susanne~C Ladd, Karin~Halina Greiser, Marc-Andre Weber,
  Jeanette Schulz-Menger, Thoralf Niendorf, Tobias Pischon, Svenja Caspers,
  Katrin Amunts, Klaus Berger, Robin Bulow, Norbert Hosten, Katrin Hegenscheid,
  Thomas Kroncke, Jakob Linseisen, Matthias Gunther, Jochen~G Hirsch, Alexander
  Kohn, Thomas Hendel, Heinz-Erich Wichmann, Borge Schmidt, Karl-Heinz Jockel,
  Wolfgang Hoffmann, Rudolf Kaaks, Maximilian~F Reiser, and Henry Volzke.
\newblock {Whole-Body MR Imaging in the German National Cohort: Rationale,
  Design, and Technical Background.}
\newblock {\em Radiology}, 277(1):206---220, 2015.

\bibitem{teismann2014}
Henning Teismann, Heike Wersching, Maren Nagel, Volker Arolt, Walter Heindel,
  Bernhard~T. Baune, Jurgen Wellmann, Hans~Werner Hense, and Klaus Berger.
\newblock {Establishing the bidirectional relationship between depression and
  subclinical arteriosclerosis - rationale, design, and characteristics of the
  BiDirect Study}.
\newblock {\em BMC Psychiatry}, 14(1):1---9, 2014.

\bibitem{kircher2019}
Tilo Kircher, Markus Wohr, Igor Nenadic, Rainer Schwarting, Gerhard Schratt,
  Judith Alferink, Carsten Culmsee, Holger Garn, Tim Hahn, Bertram
  Muller-Myhsok, Astrid Dempfle, Maik Hahmann, Andreas Jansen, Petra Pfefferle,
  Harald Renz, Marcella Rietschel, Stephanie~H. Witt, Markus Nothen, Axel Krug,
  and Udo Dannlowski.
\newblock {Neurobiology of the major psychoses: a translational perspective on
  brain structure and function—the FOR2107 consortium}.
\newblock {\em European Archives of Psychiatry and Clinical Neuroscience},
  269(8):949---962, 2019.

\bibitem{kendall2017}
Alex Kendall and Yarin Gal.
\newblock {What uncertainties do we need in Bayesian deep learning for computer
  vision?}
\newblock {\em Advances in Neural Information Processing Systems},
  2017-Decem(Nips):5575---5585, 2017.

\bibitem{liu2020}
Haitao Liu, Yew-Soon Ong, Xiaobo Shen, and Jianfei Cai.
\newblock {When Gaussian Process Meets Big Data: A Review of Scalable GPs}.
\newblock {\em IEEE Transactions on Neural Networks and Learning Systems},
  31(11):1---19, 2020.

\bibitem{tagasovska2018a}
Natasa Tagasovska and David Lopez-Paz.
\newblock {Single-Model Uncertainties for Deep Learning}.
\newblock (NeurIPS), 2018.

\bibitem{xu2017}
Qifa Xu, Kai Deng, Cuixia Jiang, Fang Sun, and Xue Huang.
\newblock {Composite quantile regression neural network with applications}.
\newblock {\em Expert Systems with Applications}, 76:129---139, 2017.

\bibitem{koenker1978}
Roger Koenker and Gilbert Bassett.
\newblock {Regression Quantiles}.
\newblock {\em Econometrica}, 46(1):33, 1978.

\bibitem{pedregosa2011}
Fabian Pedregosa, Gael Varoquaux, Alexandre Gramfort, Vincent Michel, Bertrand
  Thirion, Olivier Grisel, Mathieu Blondel, Peter Prettenhofer, Ron Weiss, and
  Vincent Dubourg.
\newblock {Scikit-learn: Machine learning in Python}.
\newblock {\em the Journal of machine Learning research}, 12:2825---2830, 2011.

\bibitem{tzourio-mazoyer2002}
N.~Tzourio-Mazoyer, B.~Landeau, D.~Papathanassiou, F.~Crivello, O.~Etard,
  N.~Delcroix, B.~Mazoyer, and M.~Joliot.
\newblock {Automated anatomical labeling of activations in SPM using a
  macroscopic anatomical parcellation of the MNI MRI single-subject brain}.
\newblock {\em NeuroImage}, 15(1):273---289, 2002.

\bibitem{halekoh2014}
Ulrich Halekoh and Sren Hjsgaard.
\newblock {A kenward-Roger approximation and parametric bootstrap methods for
  tests in linear mixed models-the R package pbkrtest}.
\newblock {\em Journal of Statistical Software}, 59(9):1---32, 2014.

\bibitem{Samek.2021}
Wojciech Samek, Grégoire Montavon, Sebastian Lapuschkin, Christopher~J.
  Anders, and Klaus-Robert Müller.
\newblock {Explaining Deep Neural Networks and Beyond: A Review of Methods and
  Applications}.
\newblock {\em Proceedings of the IEEE}, 109(3):247--278, 2021.

\end{thebibliography}

\section*{Acknowledgments}

\subsection*{Funding}
This work was funded by the German Research Foundation (DFG grants HA7070/2-2, HA7070/3, HA7070/4 to TH) and the Interdisciplinary Center for Clinical Research (IZKF) of the medical faculty of Münster (grants Dan3/012/17 to UD and MzH 3/020/20 to TH).
The analysis was conducted with data from the German National Cohort (GNC) (www.nako.de). The GNC is funded by the Federal Ministry of Education and Research (BMBF) [project funding reference numbers: 01ER1301A/B/C and 01ER1511D], the federal states and the Helmholtz Association with additional financial support by the participating universities and the institutes of the Helmholtz Association and of the Leibniz Association.  We thank all participants who took part in the GNC study and the staff in this research program.

The BiDirect Study is supported by grants from the German Ministry of Research and Education (BMBF) to the University of Muenster (grant numbers: 01ER0816 and 01ER1506).

The MACS dataset used in this work is part of the German multicenter consortium “Neurobiology of Affective Disorders. A translational perspective on brain structure and function“, funded by the German Research Foundation (Deutsche Forschungsgemeinschaft DFG; Forschungsgruppe/Research Unit FOR2107). Principal investigators (PIs) with respective areas of responsibility in the FOR2107 consortium are: Work Package WP1, FOR2107/MACS cohort and brainimaging: Tilo Kircher (speaker FOR2107; DFG grant numbers KI 588/14-1, KI 588/14-2), Udo Dannlowski (co-speaker FOR2107; DA 1151/5-1, DA 1151/5-2), Axel Krug (KR 3822/5-1, KR 3822/7-2), Igor Nenadic (NE 2254/1-2), Carsten Konrad (KO 4291/3-1). WP2, animal phenotyping: Markus Wöhr (WO 1732/4-1, WO 1732/4-2), Rainer Schwarting (SCHW 559/14-1, SCHW 559/14-2). WP3, miRNA: Gerhard Schratt (SCHR 1136/3-1, 1136/3-2). WP4, immunology, mitochondriae: Judith Alferink (AL 1145/5-2), Carsten Culmsee (CU 43/9-1, CU 43/9-2), Holger Garn (GA 545/5-1, GA 545/7-2). WP5, genetics: Marcella Rietschel (RI 908/11-1, RI 908/11-2), Markus Nöthen (NO 246/10-1, NO 246/10-2), Stephanie Witt (WI 3439/3-1, WI 3439/3-2). WP6, multi method data analytics: Andreas Jansen (JA 1890/7-1, JA 1890/7-2), Tim Hahn (HA 7070/2-2), Bertram Müller-Myhsok (MU1315/8-2), Astrid Dempfle (DE 1614/3-1, DE 1614/3-2). CP1, biobank: Petra Pfefferle (PF 784/1-1, PF 784/1-2), Harald Renz (RE 737/20-1, 737/20-2). CP2, administration. Tilo Kircher (KI 588/15-1, KI 588/17-1), Udo Dannlowski (DA 1151/6-1), Carsten Konrad (KO 4291/4-1). Data access and responsibility: All PIs take responsibility for the integrity of the respective study data and their components. All authors and coauthors had full access to all study data. The FOR2107 cohort project (WP1) was approved by the Ethics Committees of the Medical Faculties, University of Marburg (AZ: 07/14) and University of Münster (AZ: 2014-422-b-S). 
Financial support for the Beijing Normal University dataset used in this project was provided by a grant from the National Natural Science Foundation of China: 30770594 and a grant from the National High Technology Program of China (863): 2008AA02Z405.

\newpage
\section{Supplementary Material}
To assess the effects of modeling aleatory and epistemic uncertainty, we investigated PICPs for 1) the MCCQR approach modeling both aleatory and epistemic uncertainty, 2) a version of our model accounting for aleatory uncertainty only, and 3) a version of our model modeling epistemic uncertainty only. 
\begin{figure}[b]
	\centering
	\includegraphics{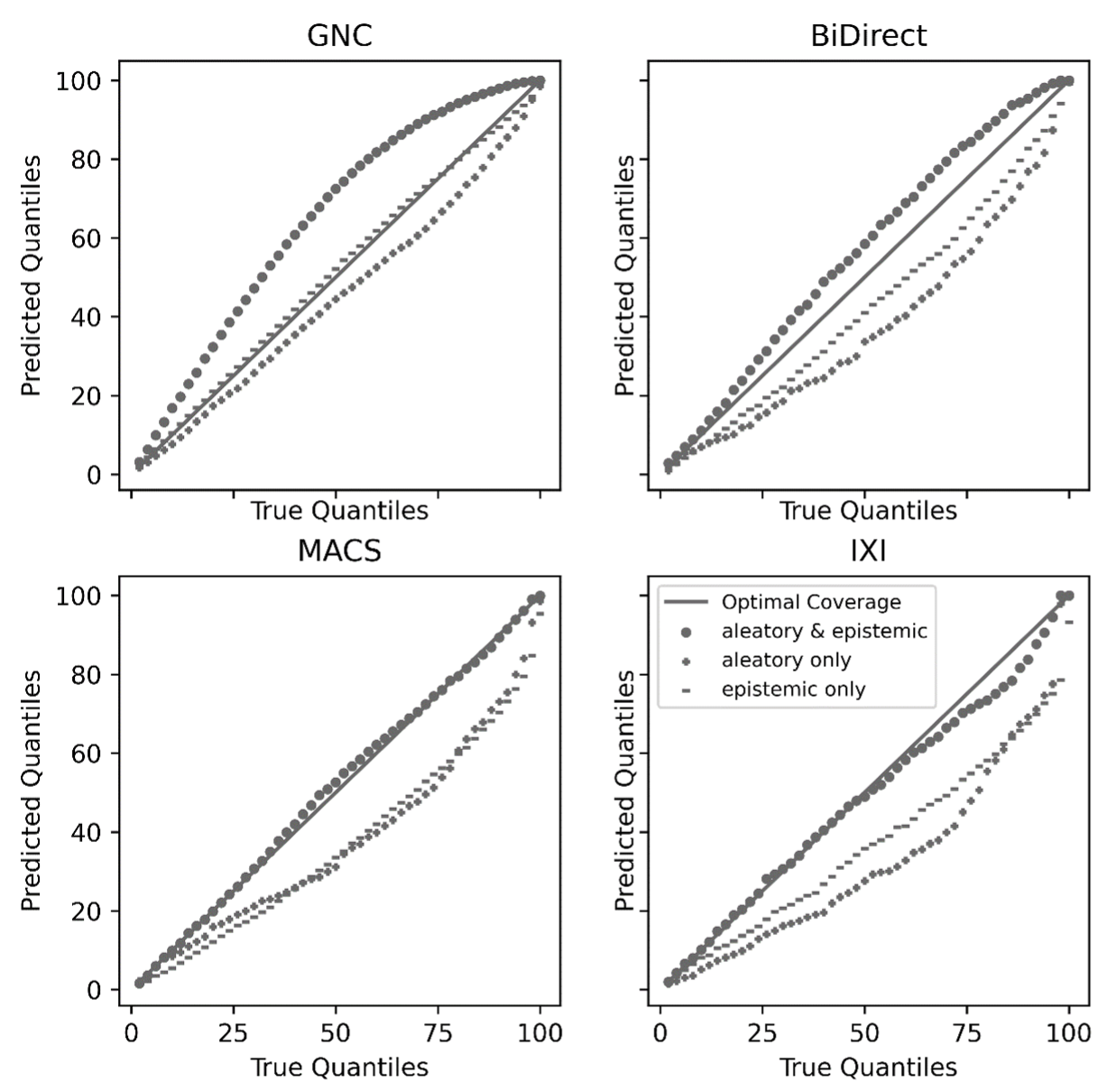}
	\renewcommand{\figurename}{Supplementary Figure}
	\renewcommand{\thefigure}{S1}
	\caption{Prediction Interval Coverage Probabilities (PICP) for leave-site-out GNC and independent validation samples (BiDirect, MACS, and IXI) for modeling 1) aleatory uncertainty only, 2) epistemic uncertainty only, and 3) both aleatory and epistemic uncertainty. Underestimation (overestimation) of uncertainty occurs, if empirical PICPs are below (above) optimal PICP as indicated by the solid line. }
\end{figure}

\begin{figure}
	\centering
	\includegraphics{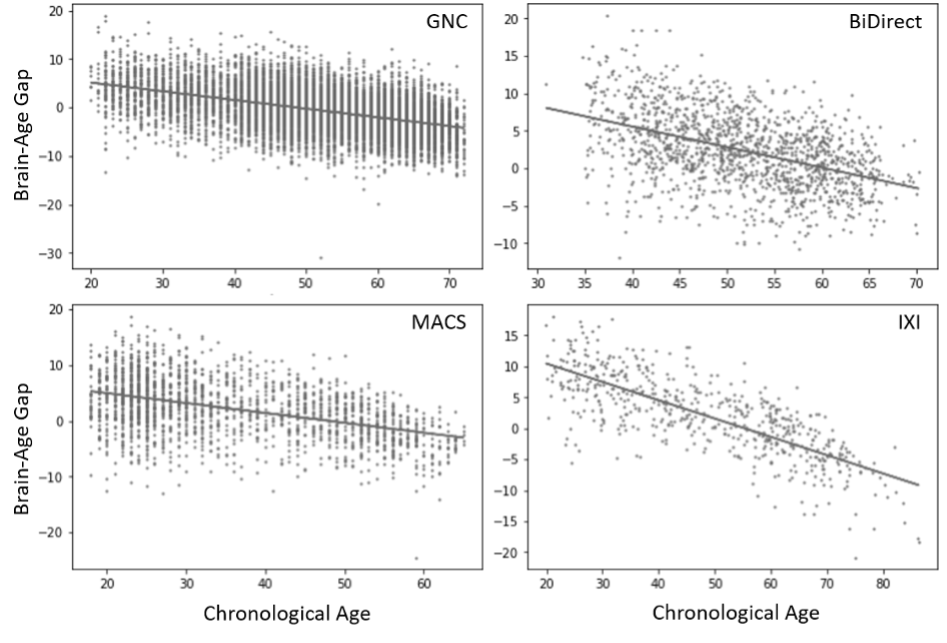}
	\renewcommand{\figurename}{Supplementary Figure}
	\renewcommand{\thefigure}{S2}
	\caption{Scatter plot showing the association between brain-age gap and chronological age for GNC, BiDirect, MACS, and IXI.}
\end{figure}

\begin{FPfigure}
	\centering
	\includegraphics{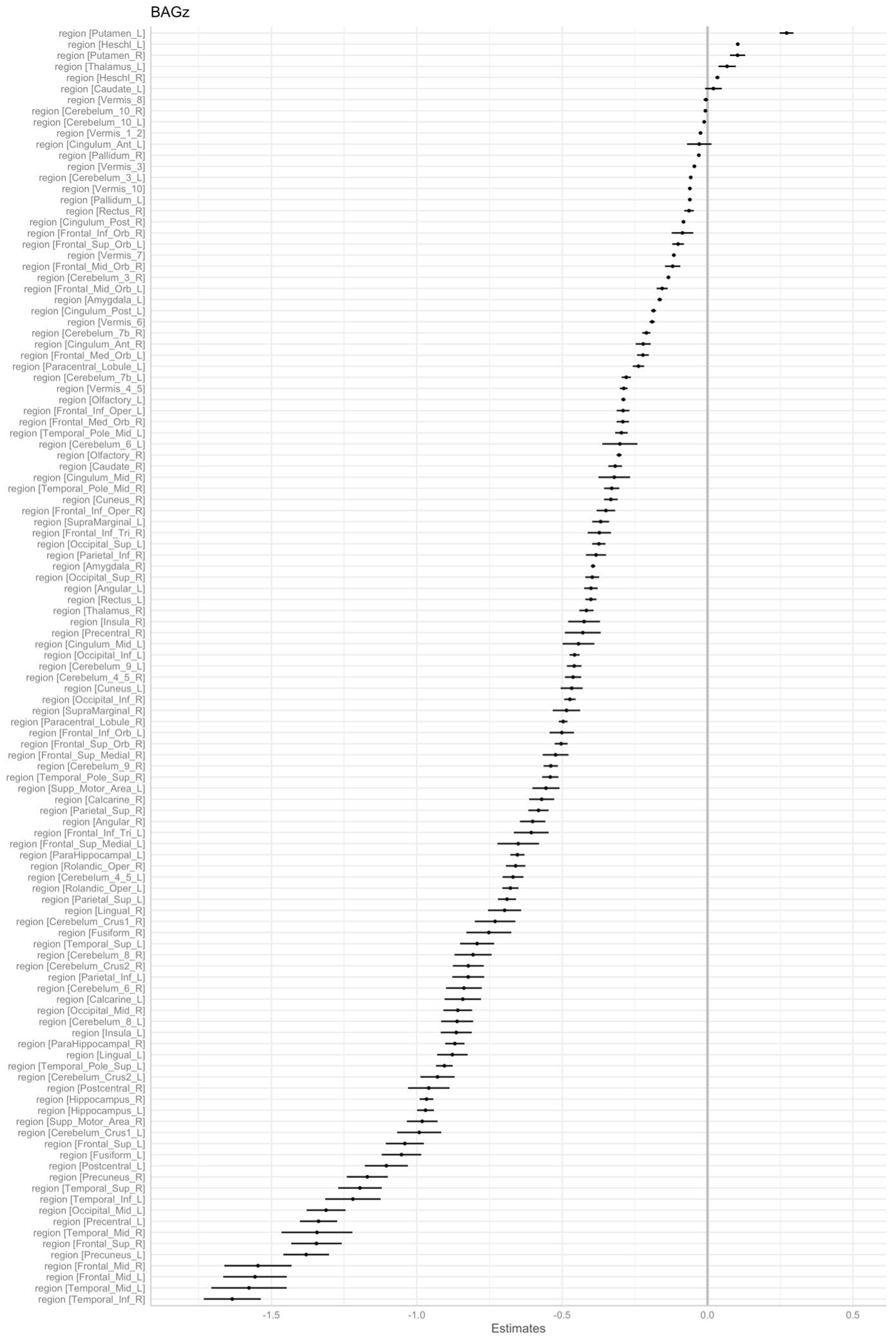}
	\renewcommand{\figurename}{Supplementary Figure}
	\renewcommand{\thefigure}{S3}
	\caption*{Supplementary Figure S3: Fixed effect estimates for AAL region of interests of the Occlusion-Sensitivity Mapping analysis. Estimates can be interpreted as difference from the whole brain, not occluded model. Error bars indicate +/- 1 standard deviation.}
\end{FPfigure}
\newpage

\begin{longtable}{l c c c}
	\hline
	& & \textbf{BAGz} & \\
	\textit{Predictors} & \textit{Estimates} & \textit{CI} & \textit{p}\\
(Intercept) & 0.97 & 0.92, 1.02 & \textbf{<0.001}\\\hline
age & -0.37 & -0.40, -0.34 & \textbf{<0.001}\\\hline
gender & 0.08 & 0.05, 0.11 & \textbf{<0.001}\\\hline
site & 0.22 & 0.18, 0.25 & \textbf{<0.001}\\\hline
ROI size & 0.41 & 0.38, 0.44 & \textbf{<0.001}\\\hline
region Amygdala\_L & -0.16 & -0.17, -0.16 & \textbf{<0.001}\\\hline
region Amygdala\_R & -0.39 & -0.40, -0.39 & \textbf{<0.001}\\\hline
region Angular\_L & -0.40 & -0.42, -0.38 & \textbf{<0.001}\\\hline
region Angular\_R & -0.60 & -0.65, -0.56 & \textbf{<0.001}\\\hline
region Calcarine\_L & -0.84 & -0.90, -0.78 & \textbf{<0.001}\\\hline
region Calcarine\_R & -0.57 & -0.61, -0.53 & \textbf{<0.001}\\\hline
region Caudate\_L & 0.02 & -0.01, 0.05 & 0.167\\\hline
region Caudate\_R & -0.32 & -0.34, -0.29 & \textbf{<0.001}\\\hline
region Cerebelum\_10\_L & -0.01 & -0.02, -0.01 & \textbf{<0.001}\\\hline
region Cerebelum\_10\_R & -0.01 & -0.01, -0.00 & 0.022\\\hline
region Cerebelum\_3\_L & -0.06 & -0.06, -0.05 & \textbf{<0.001}\\\hline
region Cerebelum\_3\_R & -0.13 & -0.14, -0.13 & \textbf{<0.001}\\\hline
region Cerebelum\_4\_5\_L & -0.67 & -0.71, -0.63 & \textbf{<0.001}\\\hline
region Cerebelum\_4\_5\_R & -0.46 & -0.49, -0.43 & \textbf{<0.001}\\\hline
region Cerebelum\_6\_L & -0.30 & -0.36, -0.24 & \textbf{<0.001}\\\hline
region Cerebelum\_6\_R & -0.84 & -0.90, -0.78 & \textbf{<0.001}\\\hline
region Cerebelum\_7b\_L & -0.28 & -0.30, -0.26 & \textbf{<0.001}\\\hline
region Cerebelum\_7b\_R & -0.21 & -0.22, -0.20 & \textbf{<0.001}\\\hline
region Cerebelum\_8\_L & -0.86 & -0.92, -0.81 & \textbf{<0.001}\\\hline
region Cerebelum\_8\_R & -0.81 & -0.87, -0.74 & \textbf{<0.001}\\\hline
region Cerebelum\_9\_L & -0.46 & -0.48, -0.43 & \textbf{<0.001}\\\hline
region Cerebelum\_9\_R & -0.54 & -0.56, -0.51 & \textbf{<0.001}\\\hline
region Cerebelum\_Crus1\_L & -0.99 & -1.07, -0.92 & \textbf{<0.001}\\\hline
region Cerebelum\_Crus1\_R & -0.73 & -0.80, -0.66 & \textbf{<0.001}\\\hline
region Cerebelum\_Crus2\_L & -0.93 & -0.99, -0.87 & \textbf{<0.001}\\\hline
region Cerebelum\_Crus2\_R & -0.82 & -0.88, -0.77 & \textbf{<0.001}\\\hline
region Cingulum\_Ant\_L & -0.03 & -0.07, 0.01 & 0.186\\\hline
region Cingulum\_Ant\_R & -0.22 & -0.25, -0.20 & \textbf{<0.001}\\\hline
region Cingulum\_Mid\_L & -0.44 & -0.50, -0.39 & \textbf{<0.001}\\\hline
region Cingulum\_Mid\_R & -0.32 & -0.38, -0.27 & \textbf{<0.001}\\\hline
region Cingulum\_Post\_L & -0.19 & -0.19, -0.18 & \textbf{<0.001}\\\hline
region Cingulum\_Post\_R & -0.08 & -0.09, -0.08 & \textbf{<0.001}\\\hline
region Cuneus\_L & -0.47 & -0.51, -0.43 & \textbf{<0.001}\\\hline
region Cuneus\_R & -0.33 & -0.36, -0.31 & \textbf{<0.001}\\\hline
region Frontal\_Inf\_Oper\_L & -0.29 & -0.31, -0.27 & \textbf{<0.001}\\\hline
region Frontal\_Inf\_Oper\_R & -0.35 & -0.38, -0.32 & \textbf{<0.001}\\\hline
region Frontal\_Inf\_Orb\_L & -0.50 & -0.54, -0.46 & \textbf{<0.001}\\\hline
region Frontal\_Inf\_Orb\_R & -0.09 & -0.12, -0.05 & \textbf{<0.001}\\\hline
region Frontal\_Inf\_Tri\_L & -0.61 & -0.67, -0.55 & \textbf{<0.001}\\\hline
region Frontal\_Inf\_Tri\_R & -0.37 & -0.41, -0.33 & \textbf{<0.001}\\\hline
region Frontal\_Med\_Orb\_L & -0.22 & -0.24, -0.20 & \textbf{<0.001}\\\hline
region Frontal\_Med\_Orb\_R & -0.29 & -0.31, -0.27 & \textbf{<0.001}\\\hline
region Frontal\_Mid\_L & -1.56 & -1.67, -1.45 & \textbf{<0.001}\\\hline
region Frontal\_Mid\_Orb\_L & -0.16 & -0.17, -0.14 & \textbf{<0.001}\\\hline
region Frontal\_Mid\_Orb\_R & -0.12 & -0.15, -0.09 & \textbf{<0.001}\\\hline
region Frontal\_Mid\_R & -1.55 & -1.66, -1.43 & \textbf{<0.001}\\\hline
region Frontal\_Sup\_L & -1.04 & -1.11, -0.98 & \textbf{<0.001}\\\hline
region Frontal\_Sup\_Medial\_L & -0.65 & -0.72, -0.58 & \textbf{<0.001}\\\hline
region Frontal\_Sup\_Medial\_R & -0.52 & -0.57, -0.48 & \textbf{<0.001}\\\hline
region Frontal\_Sup\_Orb\_L & -0.10 & -0.12, -0.08 & \textbf{<0.001}\\\hline
region Frontal\_Sup\_Orb\_R & -0.50 & -0.53, -0.48 & \textbf{<0.001}\\\hline
region Frontal\_Sup\_R & -1.35 & -1.43, -1.26 & \textbf{<0.001}\\\hline
region Fusiform\_L & -1.05 & -1.12, -0.98 & \textbf{<0.001}\\\hline
region Fusiform\_R & -0.75 & -0.83, -0.67 & \textbf{<0.001}\\\hline
region Heschl\_L & 0.10 & 0.10, 0.11 & \textbf{<0.001}\\\hline
region Heschl\_R & 0.03 & 0.03, 0.04 & \textbf{<0.001}\\\hline
region Hippocampus\_L & -0.97 & -1.00, -0.94 & \textbf{<0.001}\\\hline
region Hippocampus\_R & -0.97 & -0.99, -0.94 & \textbf{<0.001}\\\hline
region Insula\_L & -0.86 & -0.92, -0.81 & \textbf{<0.001}\\\hline
region Insula\_R & -0.42 & -0.48, -0.37 & \textbf{<0.001}\\\hline
region Lingual\_L & -0.88 & -0.93, -0.83 & \textbf{<0.001}\\\hline
region Lingual\_R & -0.70 & -0.76, -0.64 & \textbf{<0.001}\\\hline
region Occipital\_Inf\_L & -0.46 & -0.48, -0.44 & \textbf{<0.001}\\\hline
region Occipital\_Inf\_R & -0.47 & -0.49, -0.45 & \textbf{<0.001}\\\hline
region Occipital\_Mid\_L & -1.31 & -1.38, -1.25 & \textbf{<0.001}\\\hline
region Occipital\_Mid\_R & -0.86 & -0.91, -0.81 & \textbf{<0.001}\\\hline
region Occipital\_Sup\_L & -0.37 & -0.40, -0.35 & \textbf{<0.001}\\\hline
region Occipital\_Sup\_R & -0.40 & -0.42, -0.37 & \textbf{<0.001}\\\hline
region Olfactory\_L & -0.29 & -0.30, -0.28 & \textbf{<0.001}\\\hline
region Olfactory\_R & -0.30 & -0.31, -0.30 & \textbf{<0.001}\\\hline
region Pallidum\_L & -0.06 & -0.07, -0.05 & \textbf{<0.001}\\\hline
region Pallidum\_R & -0.03 & -0.04, -0.02 & \textbf{<0.001}\\\hline
region Paracentral\_Lobule\_L & -0.24 & -0.26, -0.22 & \textbf{<0.001}\\\hline
region Paracentral\_Lobule\_R & -0.50 & -0.51, -0.48 & \textbf{<0.001}\\\hline
region ParaHippocampal\_L & -0.65 & -0.68, -0.63 & \textbf{<0.001}\\\hline
region ParaHippocampal\_R & -0.87 & -0.90, -0.84 & \textbf{<0.001}\\\hline
region Parietal\_Inf\_L & -0.82 & -0.88, -0.77 & \textbf{<0.001}\\\hline
region Parietal\_Inf\_R & -0.38 & -0.42, -0.35 & \textbf{<0.001}\\\hline
region Parietal\_Sup\_L & -0.69 & -0.72, -0.66 & \textbf{<0.001}\\\hline
region Parietal\_Sup\_R & -0.58 & -0.62, -0.55 & \textbf{<0.001}\\\hline
region Postcentral\_L & -1.10 & -1.18, -1.03 & \textbf{<0.001}\\\hline
region Postcentral\_R & -0.96 & -1.03, -0.89 & \textbf{<0.001}\\\hline
region Precentral\_L & -1.34 & -1.40, -1.27 & \textbf{<0.001}\\\hline
region Precentral\_R & -0.43 & -0.49, -0.37 & \textbf{<0.001}\\\hline
region Precuneus\_L & -1.38 & -1.46, -1.30 & \textbf{<0.001}\\\hline
region Precuneus\_R & -1.17 & -1.24, -1.10 & \textbf{<0.001}\\\hline
region Putamen\_L & 0.27 & 0.25, 0.30 & \textbf{<0.001}\\\hline
region Putamen\_R & 0.10 & 0.08, 0.13 & \textbf{<0.001}\\\hline
region Rectus\_L & -0.40 & -0.42, -0.38 & \textbf{<0.001}\\\hline
region Rectus\_R & -0.06 & -0.08, -0.05 & \textbf{<0.001}\\\hline
region Rolandic\_Oper\_L & -0.68 & -0.71, -0.65 & \textbf{<0.001}\\\hline
region Rolandic\_Oper\_R & -0.66 & -0.69, -0.63 & \textbf{<0.001}\\\hline
region Supp\_Motor\_Area\_L & -0.56 & -0.60, -0.51 & \textbf{<0.001}\\\hline
region Supp\_Motor\_Area\_R & -0.98 & -1.03, -0.93 & \textbf{<0.001}\\\hline
region SupraMarginal\_L & -0.37 & -0.40, -0.34 & \textbf{<0.001}\\\hline
region SupraMarginal\_R & -0.49 & -0.53, -0.44 & \textbf{<0.001}\\\hline
region Temporal\_Inf\_L & -1.22 & -1.32, -1.12 & \textbf{<0.001}\\\hline
region Temporal\_Inf\_R & -1.63 & -1.73, -1.54 & \textbf{<0.001}\\\hline
region Temporal\_Mid\_L & -1.58 & -1.71, -1.45 & \textbf{<0.001}\\\hline
region Temporal\_Mid\_R & -1.34 & -1.47, -1.22 & \textbf{<0.001}\\\hline
region Temporal\_Pole\_Mid\_L & -0.30 & -0.32, -0.27 & \textbf{<0.001}\\\hline
region Temporal\_Pole\_Mid\_R & -0.33 & -0.36, -0.30 & \textbf{<0.001}\\\hline
region Temporal\_Pole\_Sup\_L & -0.91 & -0.93, -0.88 & \textbf{<0.001}\\\hline
region Temporal\_Pole\_Sup\_R & -0.54 & -0.57, -0.51 & \textbf{<0.001}\\\hline
region Temporal\_Sup\_L & -0.79 & -0.85, -0.73 & \textbf{<0.001}\\\hline
region Temporal\_Sup\_R & -1.20 & -1.27, -1.12 & \textbf{<0.001}\\\hline
region Thalamus\_L & 0.07 & 0.04, 0.10 & \textbf{<0.001}\\\hline
region Thalamus\_R & -0.42 & -0.44, -0.39 & \textbf{<0.001}\\\hline
region Vermis\_1\_2 & -0.02 & -0.03, -0.02 & \textbf{<0.001}\\\hline
region Vermis\_10 & -0.06 & -0.07, -0.05 & \textbf{<0.001}\\\hline
region Vermis\_3 & -0.05 & -0.05, -0.04 & \textbf{<0.001}\\\hline
region Vermis\_4\_5 & -0.29 & -0.30, -0.27 & \textbf{<0.001}\\\hline
region Vermis\_6 & -0.19 & -0.20, -0.18 & \textbf{<0.001}\\\hline
region Vermis\_7 & -0.12 & -0.12, -0.11 & \textbf{<0.001}\\\hline
region Vermis\_8 & -0.01 & -0.01, 0.00 & 0.154\\\hline
& & & \\
\textbf{Random Effects} & & & \\
$\sigma^2$ & 0.01 & &\\
$\tau_{00\text{ Subject ID}}$ & 0.52 & & \\
ICC & 0.98 & & \\
$N_\text{Subject ID}$ & 1986 & &\\\hline
Observations & 232362 & &  \\
Marginal $R^2$ / Conditional $R^2$ & 0.291 / 0.984 & & \\
	\caption*{Supplementary Table S1: Results of the General Multi-Level Model for Occlusion-Sensitivity Mapping. CI = confidence intervals, $\sigma^2$ = variance of random effect, ICC = intra class correlation coefficient, N = number of samples, $R^2$ = explained variance.)}
\end{longtable}

\begin{table}[h]
	\centering
	\begin{tabular}{|l|l|l|l|l|l|l|l|}
		\hline
		\textbf{Sample} & \textbf{Group} &\textbf{N} & \textbf{N Males} & \textbf{Age Mean} & \textbf{Age Std.} & \textbf{Age Min.} & \textbf{Age Max.}\\\hline\hline
		\textbf{GNC} & \textbf{Full Sample} & \textbf{10,691} & \textbf{5,485} & \textbf{51.79} & \textbf{11.37} & \textbf{20.00} & \textbf{72.00}\\\hline
		\textbf{BiDirect} & \textbf{Full Sample} & \textbf{1,460} & \textbf{811} & \textbf{51.37} & \textbf{8.15} & \textbf{30.91} & \textbf{70.25}\\\hline
		 & MDD & 719 & 438 & 49.39 & 7.40 & 30.91 & 67.09\\\hline
		 & Cardiac Event & 53 & 41 & 56.96 & 5.97 & 43.18 & 66.54\\\hline
		 & Population Sample & 688 & 361 & 53.01 & 8.48 & 35.19 & 70.25\\\hline
		 \textbf{MACS} & \textbf{Full Sample} & \textbf{1,986} & \textbf{1,259} & \textbf{35.87} & \textbf{13.03} & \textbf{18.00} & \textbf{65.00}\\\hline
		 & HC & 924 & 595 & 34.16 & 12.88 & 18.00 & 65.00\\\hline
		 & MDD & 822 & 537 & 36.51 & 13.18 & 18.00 & 65.00\\\hline
		 & BD & 131 & 71 & 41.76 & 11.73 & 20.00 & 64.00\\\hline
		 & SZ & 66 & 35 & 38.03 & 11.08 & 18.00 & 57.00\\\hline
		 & SA & 43 & 25 & 38.91 & 13.15 & 18.00 & 63.00\\\hline
		 \textbf{IXI} & \textbf{HC} & \textbf{561} & \textbf{311} & \textbf{48.62} & \textbf{16.49} & \textbf{19.98} & \textbf{86.32}\\\hline
		 \textbf{Beijing Normal University} & \textbf{HC} & \textbf{179} &\textbf{107}&\textbf{21.25}&\textbf{1.92}&\textbf{18.00}&\textbf{28.00}\\\hline
	\end{tabular}
	\renewcommand{\tablename}{Supplementary Table}
	\renewcommand{\thetable}{S2}
	\caption{Overview of sample and subsample distributions. Std.: Standard Deviation, Min.: Minimum, Max.: Maximum, HC: Healthy Controls, MDD: Major Depressive Disorder, BD: Bipolar Disorder, SZ: Schizophrenia, SA: Schizoaffective Disorder)}
\end{table}
\end{document}